\documentclass[11pt]{article}

\usepackage[preprint]{acl}

\usepackage{times}
\usepackage{latexsym}
\usepackage[T1]{fontenc}
\usepackage[utf8]{inputenc}
\usepackage{microtype}
\usepackage{inconsolata}

\usepackage{amsmath}
\usepackage{amssymb}
\usepackage{amsthm}
\usepackage{bm}
\usepackage{booktabs}
\usepackage{multirow}
\usepackage{xcolor}
\usepackage{graphicx}
\usepackage{cuted}
\usepackage{capt-of}
\usepackage{tikz}
\usetikzlibrary{arrows.meta,positioning,fit,backgrounds,calc,patterns}

\title{AlignAtt4LLM: Fast AlignAtt for Decoder-Only LLMs
\\ at IWSLT 2026 Simultaneous Speech Translation Task}
\author{
Quentin Fuxa \\
Independent Researcher \\
\texttt{quentin.fuxa@gmail.com}
\And
Dominik Mach\'a\v{c}ek \\
Charles University, MFF, \'{U}FAL \\
\& University of Edinburgh \\
\texttt{machacek@ufal.mff.cuni.cz}
}
\usepackage{cleveref} 

\def\Aref#1{Appendix~\ref{#1}}
\def\AAName{AlignAtt4LLM}
\newcommand{\offlinecell}[1]{\textcolor{gray}{\emph{#1}}}

\begin{document}
\maketitle

\begin{abstract}
We describe \AAName{}, an IWSLT 2026 simultaneous speech translation system
for English to German, Italian, and Chinese. The system
is a synchronous cascade: Qwen3-ASR with forced alignment produces an
incrementally updated source transcript, and Gemma-4 E4B-it
translates that prefix under an MT-side AlignAtt policy.

To our knowledge, this is the first application of AlignAtt to a decoder-only
LLM, where the encoder-decoder cross-attention used by earlier AlignAtt
systems is absent. We recover a usable policy by proposing (1) an explicit source
span in the prompt, (2) offline selection of translation-specific
alignment heads, (3) selective \emph{qk-fast} replay of the draft-to-source
attention block, and (4) runtime query/key capture that preserves
model outputs bit-identically.

On the IWSLT 2026 development set, \AAName{} outperforms the supplied baselines for the European target languages, English to German and English to Italian, in both the low-latency regime around 2 seconds and the high-latency regime below 4 seconds CU-LongYAAL. Results for English to Chinese are more mixed, but the method is not tied to Gemma-4: because AlignAtt4LLM only requires a deterministic prompt layout, calibrated attention heads, and query/key capture, the same policy can be reapplied to stronger translation-focused decoder-only MT backbones for non-European target languages.
\end{abstract}

\section{Introduction}
\label{sec:intro}

This paper describes the \AAName{} IWSLT 2026 submission for
English to German, Italian, and Chinese simultaneous
speech translation \citep{adelani-etal-2026-iwslt}. The system is a
synchronous cascade: Qwen3-ASR with the
Qwen3 forced aligner \citep{qwen3asr26} produces an incrementally updated
source transcript with word end times, and Gemma-4 E4B-it \citep{gemma4card26}
translates the current source text under an MT-side AlignAtt policy.

The central contribution is to make an offline decoder-only LLM usable in simultaneous mode. Standard AlignAtt reads encoder-decoder cross-attention, but decoder-only LLMs have no such cross-attention. We instead expose the source transcript as
a known prompt span, select a small set of translation-specific self-attention
heads offline, and accept only draft prefixes whose reconstructed attention
signal remains within the currently available source frontier.

The second contribution is making the implementation fast enough for
computational-aware (CA) evaluation. Both ASR and MT are served through vLLM \citep{vllm23},
which gives the cascade one high-throughput inference stack and hot model reuse
across chunks. On the MT side, this also means the policy cannot inspect a
Python-visible attention matrix: self-attention is hidden inside fused kernels.
We therefore capture the exact query/key tensors
consumed by the deployed Gemma attention module and replay only the
draft-to-source block needed by AlignAtt.

Our implementation is available at \url{https://github.com/QuentinFuxa/AlignAtt4LLM}.

\Cref{sec:background} describes the context of the task and prior work,
\Cref{sec:system} describes the cascade,
\Cref{sec:method} details the MT-side AlignAtt realization, and
\Cref{sec:results} reports the evaluation.

\section{Background}
\label{sec:background}

\textbf{AlignAtt} \citep{alignatt23,streamatt24} is a simultaneous policy that
lets an offline sequence-to-sequence model operate in simultaneous mode.
At each generation step, the policy derives a source position for the next
target token from attention and stops generation when that position crosses the
accessible-source frontier. Earlier AlignAtt systems obtained this signal from
encoder-decoder cross-attention. AlignAtt is currently recognized as state of the art, it has been used in e.g.\ by the top-performing IWSLT 2025 system \citep{cuniiwslt25}.

\textbf{Offline models in simultaneous mode.} Repurposing offline models for
simultaneous translation is a highly promising approach because it allows reusing high-quality,
multilingual, instruction-following models without training dedicated simultaneous
models for every language direction. However, the disadvantage is that an offline model is not
trained to decide when a partial source prefix is sufficient, and naive
re-translation can flicker or hallucinate on incomplete input. Simultaneous policies
such as AlignAtt provide the missing commit decision while preserving the
 quality and flexibility of the offline model \citep{alignatt23,cuniiwslt25}.

\textbf{Decoder-only LLMs} have recently become central in high-quality
text-to-text MT systems \citep{kocmi-etal-2024-findings,kocmi-etal-2025-findings}.
Prior simultaneous systems have already used instruction-tuned decoder-only
LLMs by prompting them with the translation direction, current source prefix,
and already emitted target prefix \citep{cuniiwslt25}. Long context,
instruction following, and in-context examples make this family attractive for
robust MT, but previous decoder-only simultaneous implementations used
LocalAgreement \citep{polak-etal-2022-cuni,polak23_interspeech} rather than
AlignAtt. LocalAgreement is a strong simultaneous policy, but it does not
take advantage of model-internal attention and generally introduces more latency than AlignAtt \citep{cuniiwslt25}.

\textbf{Translation-specific attention heads.} \citet{tokenheads26} show that
multilingual decoder-only LLMs contain a sparse set of heads whose attention
argmaxes track source-target word alignments. This suggests a decoder-only
AlignAtt path, but only if two runtime conditions are met: the source span must
be identifiable in the prompt, and the policy-visible attention rows
must be reconstructed from the same tensors used by the deployed inference
engine. Our proposed method supplies both conditions.

\subsection{IWSLT 2026 Simultaneous Task}

\textbf{Simulation environment.} The IWSLT 2026 Simultaneous Task proposes the
Simulstream toolkit \citep{simulstream25} for simulating simultaneous
translation sessions. We follow the long-form, unsegmented task setup, which
allows re-translation at the evaluation interface, but \AAName{} itself emits
append-only incremental output, the mode preferred in real-world deployments where re-translation flicker could disrupt readers following the live text.

\textbf{Development data.} We use the official IWSLT 2026 MCIF development
set.
We report translation quality with BLEU
\citep{papineni-etal-2002-bleu}, chrF \citep{popovic-2015-chrf}, and
XCOMET-XL \citep{xcometxl24}. Translation latency is reported with LongYAAL
\citep{longyaal25} in both computational-unaware (CU) and
computational-aware (CA) variants.

\section{\AAName{} System Overview}
\label{sec:system}

\paragraph{ASR.}
The speech component of the cascade uses two upstream Qwen components:
Qwen3-ASR-1.7B transcribes the live audio tail, and Qwen3-ForcedAligner-0.6B
assigns word-level start and end times to that transcript \citep{qwen3asr26}.
At every chunk boundary, the cascade re-transcribes the current live utterance
tail and invokes the forced aligner with timestamp output enabled; the word
times are therefore produced online inside our simulation run, not added later
as an offline post-processing step. Adjacent ASR hypotheses are stabilized by a
longest-common-prefix commit rule up to sentence-final punctuation, while the
remaining live tail is allowed to change on the next chunk.
We retain this dedicated aligner because it performed better than all the alternative candidates summarized in \Aref{app:appendix-asr}.

\paragraph{MT.}
The translation component uses Gemma-4 E4B-it \citep{gemma4card26} served through
vLLM \citep{vllm23}.
With every source update, Gemma receives the current transcript prefix, the already accepted
target prefix, and a fixed translation instruction, then greedily generates a
draft continuation of at most 16 new tokens.
AlignAtt accepts the longest draft prefix whose
alignment signal stays on the currently accessible side of the source frontier.

\paragraph{Synchronization.}
The cascade is synchronous and chunk-based. Each chunk first updates the ASR
transcript prefix and then launches one MT request. Source words are considered
accessible only once their end time is older than a hold-back margin; in the
official IWSLT runs, this margin is 0 ms, so a word becomes accessible as
soon as its aligned end time is observed (see Section~\ref{sec:asr-tail-analysis} for a reliability analysis suggesting a conservative 250\,ms tail hold-back). Both regimes also delay the
first MT emission until 2 seconds of source audio have accumulated.

This
conservative synchronous regime cleanly separates policy behavior from scheduler overlap
and defines the CA setting used in our experiments.
\Cref{fig:system} summarizes this chunk-synchronous runtime path. Appendix~\ref{app:sync} show alternative synchronization modes.

\paragraph{Latency-quality parameters.}
We keep two official latency regimes. The low-latency submission uses
$\Delta_{\mathrm{chunk}}=850$ ms and the high-latency submission uses
$\Delta_{\mathrm{chunk}}=1500$ ms. Both use the same ASR backend, Gemma-4 MT
backbone, and MT-side AlignAtt realization. They also use
the same per-direction top-8 MT head sets, a width-7 source-axis median filter
applied before taking the source-side argmax, and a one-source-token border
margin $b=1$.


\input{figures/system_cascade_step.tex}

\section{AlignAtt for Decoder-Only LLMs}
\label{sec:method}

This section focuses on the method that enables AlignAtt for decoder-only LLMs.

\subsection{Prompt Layout}

Standard AlignAtt does not directly apply because a decoder-only LLM has no
decoder cross-attention. We therefore make the source span explicit in the
causal prompt. At chunk $k$, the serialized chat prompt is a concatenation of
the system prompt, live transcript prefix, translation instruction, accepted
translation prefix, and current draft:
\begin{equation}
\mathbf{p}^{(k)} =
\left[\mathbf{p}^{\mathrm{sys}}
\Vert \mathbf{s}^{(k)}
\Vert \mathbf{p}^{\mathrm{instr}}
\Vert \mathbf{y}_{1:m_k}
\Vert \widetilde{\mathbf{y}}^{(k)}\right],
\label{eq:mt-prompt}
\end{equation}
where $\mathbf{s}^{(k)}$ is the live transcript prefix returned by ASR,
$\mathbf{y}_{1:m_k}$ is the already accepted translation prefix, and
$\widetilde{\mathbf{y}}^{(k)}$ is the draft proposed by Gemma-4 at the current
step. The transcript prefix is therefore a contiguous prompt span with a known
map $\phi^{(k)}$ from source-word indices to prompt positions.

\begin{figure*}[t]
\centering
\resizebox{\linewidth}{!}{%
\begin{tikzpicture}[
  x=1cm, y=1cm,
  >={Latex[length=2.4mm]},
  every node/.style={font=\small, inner sep=0pt},
  panel/.style={draw=black!40, rounded corners=3pt, line width=0.6pt, fill=white},
  title/.style={font=\scriptsize\bfseries, text=black!85, align=left, inner sep=0pt},
  sub/.style={font=\scriptsize, text=black!60, align=left, inner sep=0pt},
  tny/.style={font=\tiny, text=black!55, align=center, inner sep=0pt},
  decbox/.style={draw=green!45!black, fill=green!10, rounded corners=2pt,
                 line width=0.55pt, inner sep=2pt, font=\scriptsize, align=center},
  caption/.style={font=\tiny, text=green!28!black, align=center, inner sep=0pt},
  flow/.style={->, draw=black!72, line width=0.85pt},
  peakflow/.style={->, draw=teal!55!black, line width=0.7pt, dashed},
  srcacc/.style={draw=blue!50!black, fill=blue!16, line width=0.4pt},
  srcinacc/.style={draw=blue!30!black, fill=blue!5, line width=0.4pt},
  prompt/.style={draw=black!35, fill=black!7, line width=0.35pt},
  tgtpref/.style={draw=green!45!black, fill=green!16, line width=0.35pt},
  draft/.style={draw=orange!75!black, fill=orange!18, line width=0.35pt},
]

  \def\aL{0.00}\def\aR{8.40}\def\aB{2.30}\def\aT{6.20}
  \draw[panel] (\aL,\aB) rectangle (\aR,\aT);

  \node[title, anchor=north west] at (\aL+0.25,\aT-0.15)
    {a.~Standard encoder-decoder AlignAtt};
  \node[sub, anchor=north west, text width=7.90cm] at (\aL+0.25,\aT-0.52)
    {decoder cross-attention already returns a source-only row; the policy
     reads it directly.};

  \node[decbox, minimum width=1.80cm, minimum height=0.75cm]
    (dec) at (1.50,4.40) {decoder state\\for $y_t$};

  \def\aHx{3.00}        
  \def\aHy{4.00}        
  \def\aHw{0.46}        
  \def\aHg{0.12}        
  \def\aHs{0.58}        
  \foreach \i/\h in {0/0.12, 1/0.28, 2/0.55, 3/0.82, 4/0.55, 5/0.30, 6/0.14, 7/0.08} {
    \pgfmathsetmacro\x{\aHx + \aHs*\i}
    \filldraw[fill=teal!35, draw=teal!45!black, line width=0.25pt]
      (\x,\aHy) rectangle (\x+\aHw,\aHy+\h);
  }
  \pgfmathsetmacro\aHxR{\aHx + \aHs*7 + \aHw}   
  \node[tny, anchor=south] at ({(\aHx+\aHxR)/2}, {\aHy+0.95})
    {cross-attention row, source-normalized};

  \draw[flow] (dec.east) -- (\aHx-0.10,4.40);

  \def\aSyB{2.60}\def\aSyT{3.20}
  \pgfmathsetmacro\aPeakX{\aHx + \aHs*3 + \aHw/2}
  \pgfmathsetmacro\aFx{\aHx + \aHs*3 + \aHw + \aHg/2}
  \filldraw[srcacc] (\aHx,\aSyB) rectangle (\aFx,\aSyT);
  \filldraw[srcinacc] (\aFx,\aSyB) rectangle (\aHxR,\aSyT);
  \draw[black!65, dashed, line width=0.5pt] (\aFx,\aSyB-0.20) -- (\aFx,\aHy-0.05);

  \node[tny, text=blue!55!black] at ({(\aHx+\aFx)/2}, {(\aSyB+\aSyT)/2})
    {accessible source};
  \node[tny, text=blue!40!black] at ({(\aFx+\aHxR)/2}, {(\aSyB+\aSyT)/2})
    {future source};
  \node[tny, anchor=north] at (\aFx,\aSyB-0.05) {frontier};

  \draw[peakflow] (\aPeakX,\aHy-0.05) -- (\aPeakX,\aSyT+0.05);

  \def\bL{8.60}\def\bR{17.00}\def\bB{2.30}\def\bT{6.20}
  \draw[panel] (\bL,\bB) rectangle (\bR,\bT);

  \node[title, anchor=north west] at (\bL+0.25,\bT-0.15)
    {b.~Decoder-only LLM with explicit source span};
  \node[sub, anchor=north west, text width=7.90cm] at (\bL+0.25,\bT-0.52)
    {self-attention mixes the full causal prompt; we mark the source span,
     then keep only the draft-to-source slice.};

  \def\bPx{8.85}\def\bPxR{16.75}
  \def\bPyB{4.35}\def\bPyT{4.85}
  \def\bP1{0.80}\def\bSacc{2.40}\def\bSinacc{1.00}\def\bInstr{0.80}
  \def\bYpref{1.50}\def\bDraft{1.40}
  \pgfmathsetmacro\bx{\bPx}
  \filldraw[prompt] (\bx,\bPyB) rectangle ({\bx+\bP1},\bPyT);
  \node[tny] at ({\bx+\bP1/2},{(\bPyB+\bPyT)/2}) {$\mathcal{P}_1$};
  \pgfmathsetmacro\bx{\bPx + \bP1}
  \filldraw[srcacc] (\bx,\bPyB) rectangle ({\bx+\bSacc},\bPyT);
  \node[tny, text=blue!55!black] at ({\bx+\bSacc/2},{(\bPyB+\bPyT)/2})
    {$\mathcal{S}^{(k)}_{\mathrm{acc}}$};
  \pgfmathsetmacro\bx{\bPx + \bP1 + \bSacc}
  \pgfmathsetmacro\bFx{\bx}       
  \filldraw[srcinacc] (\bx,\bPyB) rectangle ({\bx+\bSinacc},\bPyT);
  \node[tny, text=blue!40!black] at ({\bx+\bSinacc/2},{(\bPyB+\bPyT)/2})
    {$\mathcal{S}^{(k)}_{\mathrm{inacc}}$};
  \pgfmathsetmacro\bx{\bPx + \bP1 + \bSacc + \bSinacc}
  \filldraw[prompt] (\bx,\bPyB) rectangle ({\bx+\bInstr},\bPyT);
  \node[tny] at ({\bx+\bInstr/2},{(\bPyB+\bPyT)/2}) {instr};
  \pgfmathsetmacro\bx{\bPx + \bP1 + \bSacc + \bSinacc + \bInstr}
  \filldraw[tgtpref] (\bx,\bPyB) rectangle ({\bx+\bYpref},\bPyT);
  \node[tny, text=green!35!black] at ({\bx+\bYpref/2},{(\bPyB+\bPyT)/2})
    {$\mathbf{y}_{1:m_k}$};
  \pgfmathsetmacro\bx{\bPx + \bP1 + \bSacc + \bSinacc + \bInstr + \bYpref}
  \filldraw[draft] (\bx,\bPyB) rectangle ({\bx+\bDraft},\bPyT);
  \node[tny, text=orange!65!black] at ({\bx+\bDraft/2},{(\bPyB+\bPyT)/2})
    {$\widetilde{\mathbf{y}}^{(k)}$};

  \node[tny, anchor=south] at ({(\bPx+\bPxR)/2}, \bPyT+0.05)
    {explicit prompt partition};

  \draw[black!65, dashed, line width=0.5pt] (\bFx,2.70) -- (\bFx,\bPyT+0.05);

  \def\bHx{8.85}\def\bHxR{16.75}\def\bHyB{2.70}
  \def\bHw{0.45}\def\bHs{0.60}     
  \def\bHxStart{9.10}
  \foreach \i/\h/\m in {
    0/0.10/0, 1/0.32/1, 2/0.60/1, 3/0.85/1, 4/0.60/1, 5/0.30/1,
    6/0.14/1, 7/0.10/0, 8/0.12/0, 9/0.18/0, 10/0.20/0, 11/0.08/0
  } {
    \pgfmathsetmacro\x{\bHxStart + \bHs*\i}
    \ifnum\m=1
      \filldraw[fill=teal!35, draw=teal!45!black, line width=0.25pt]
        (\x,\bHyB) rectangle (\x+\bHw,\bHyB+\h);
    \else
      \filldraw[fill=black!15, draw=black!30, line width=0.22pt]
        (\x,\bHyB) rectangle (\x+\bHw,\bHyB+\h);
    \fi
  }
  \pgfmathsetmacro\bPeakX{\bHxStart + \bHs*3 + \bHw/2}

  \node[tny, anchor=north] at ({(\bHx+\bHxR)/2}, \bHyB-0.05)
    {full self-attention row; teal = source columns, gray = non-source};

  \draw[peakflow] (\bPeakX, \bHyB+0.90) -- (\bPeakX, \bPyB-0.05);

\end{tikzpicture}
}%
\caption{\textbf{How AlignAtt changes substrate between encoder-decoder and
decoder-only models.} \emph{(a)} In the original encoder-decoder setting,
the decoder already exposes a source-only cross-attention row, so the policy
can gate tokens directly against the accessible source frontier: if the peak
of the row falls on the accessible side, the draft token is accepted.
\emph{(b)} In our decoder-only MT setting, source and target history share
one causal prompt. We therefore mark the source span explicitly in prompt
space, isolate translation-specific heads, and reconstruct only the
draft-to-source slice needed by the policy; the decision rule is otherwise
the same as in (a).}
\label{fig:alignatt-substrates}
\end{figure*}
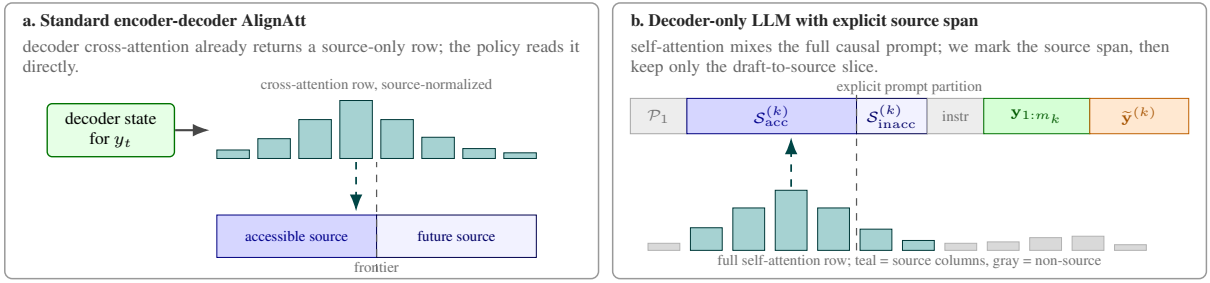

Given this prompt layout, \Cref{fig:alignatt-substrates} summarizes the
conceptual shift: the policy intuition remains the same as in encoder-decoder
AlignAtt, but the alignment signal must now be recovered from a
prompt-structured self-attention substrate rather than read directly from
cross-attention.

\subsection{Selection of Alignment Heads}

Not all self-attention heads carry a useful translation signal.
Following \citet{tokenheads26}, we operationalize this
with a two-stage offline calibration procedure under the exact prompt layout of
Eq.~\eqref{eq:mt-prompt}: GPT-5-mini \citep{openai2025gpt5developers} first
provides word-level source-target alignments on held-out parallel text, then
every Gemma head is scored with Translation Score (TS), the aligned-token
argmax accuracy used by \citet{tokenheads26}, on those aligned examples. We
retain the top $k=8$ heads per language pair. This head set is
fixed at inference time and is the only part of the policy that depends on
offline calibration.
The retained MT heads are sparse and
concentrated in a limited late-depth region of the backbone rather than being
spread uniformly; Appendix~\ref{sec:appendix-alignatt} visualizes that pattern.

The policy only reads the source slice of each draft attention row,
$\mathbf{A}^{(\ell,h)}_{t,\phi^{(k)}(s)}$, but unlike encoder-decoder
cross-attention, this slice is not source-normalized. Off-source mass on the
accepted target prefix, prompt template, and speculative suffix is therefore
structural rather than implementation noise. On the 78 text-only qualitative
probes used for diagnostics, drafted units allocate on average 9\%
to accessible source tokens, 8\% to inaccessible source tokens, 81\% to
non-source prompt positions, and 2\% to the speculative suffix; among units
that the policy actually accepts, the inaccessible-source share drops from 8\%
to 1\%.

\subsection{Selective Reconstruction for the Policy}

\begin{figure*}[t]
\centering
\resizebox{\linewidth}{!}{%
\begin{tikzpicture}[
  x=1cm, y=1cm,
  >={Latex[length=2.6mm]},
  every node/.style={font=\small},
  panel/.style={draw=black!45, rounded corners=3pt, line width=0.6pt, fill=white},
  title/.style={font=\bfseries\small, text=black!85},
  subtitle/.style={font=\scriptsize, text=black!60, align=left},
  stagearrow/.style={->, line width=1.1pt, draw=black!75},
  guide/.style={draw=black!25, line width=0.35pt, dashed},
  kernelbox/.style={draw=black!35, dashed, fill=black!3, line width=0.6pt},
  captureband/.style={fill=orange!20, draw=orange!75!black, line width=0.9pt},
  captureline/.style={draw=orange!75!black, line width=0.9pt, rounded corners=1pt},
  reconbox/.style={draw=teal!65!black, fill=teal!15, line width=1.1pt, rounded corners=1.5pt},
  partP/.style={draw=black!30, fill=black!7, line width=0.3pt},
  partPy/.style={draw=green!35!black, fill=green!18, line width=0.3pt},
  partSa/.style={draw=blue!40, fill=blue!18, line width=0.3pt},
  partSi/.style={draw=blue!25, fill=blue!6, line width=0.3pt},
  partD/.style={draw=purple!45, fill=purple!14, line width=0.3pt},
  notrecon/.style={pattern=north east lines, pattern color=black!25, draw=black!20, line width=0.2pt},
  labelnote/.style={font=\scriptsize, text=black!70, align=center},
  mathnote/.style={font=\scriptsize\itshape, text=black!60, align=center},
]

  \draw[panel] (-1.55, 0.35) rectangle (9.3, 6.85);
  \node[title, anchor=north west] at (-1.30, 6.75) {1.~Fused attention forward};

  \def\mL{1.70}\def\mR{7.70}\def\mB{0.55}\def\mT{5.55}
  \def\cPa{2.10}
  \def\cSa{3.70}
  \def\cSi{4.60}
  \def\cPi{5.30}
  \def\cPy{6.90}
  \def\rPa{4.95}
  \def\rSa{3.75}
  \def\rSi{2.95}
  \def\rPi{2.40}
  \def\rPy{1.25}

  \draw[kernelbox] (\mL, \mB) rectangle (\mR, \mT);
  \node[mathnote] at ({(\mL+\mR)/2}, 4.40)
    {$\mathbf{A}^{(\ell,h)}=\mathrm{softmax}\!\bigl(\gamma\,\mathbf{Q}\mathbf{K}^{\!\top}+\mathbf{M}\bigr)$};
  \node[labelnote] at ({(\mL+\mR)/2}, 3.95) {$n{\times}n$, fused, never materialized};

  \foreach \x in {\cPa,\cSa,\cSi,\cPi,\cPy} \draw[guide] (\x,\mB) -- (\x,\mT);
  \foreach \y in {\rPa,\rSa,\rSi,\rPi,\rPy} \draw[guide] (\mL,\y) -- (\mR,\y);

  \draw[notrecon] (\mL,  \mB) rectangle (\cPa,  \rPy);
  \draw[notrecon] (\cSi, \mB) rectangle (\mR,   \rPy);

  \draw[reconbox] (\cPa, \mB) rectangle (\cSi, \rPy);
  \node[font=\scriptsize\bfseries, text=teal!55!black]
    at ({(\cPa+\cSi)/2}, {(\mB+\rPy)/2 + 0.07})
    {$\widehat{\mathbf{A}}_{\mathcal{D},\mathcal{S}^{(k)}}$};
  \node[font=\tiny, text=teal!55!black]
    at ({(\cPa+\cSi)/2}, {(\mB+\rPy)/2 - 0.17})
    {$L_k{\times}N_k$, only block used};

  \def\csB{5.68}\def\csT{6.00}
  \draw[partP ] (\mL  ,\csB) rectangle (\cPa ,\csT) node[midway,font=\tiny]{$\mathcal{P}_1$};
  \draw[partSa] (\cPa ,\csB) rectangle (\cSa ,\csT) node[midway,font=\tiny]{$\mathcal{S}^{(k)}_{\mathrm{acc}}$};
  \draw[partSi] (\cSa ,\csB) rectangle (\cSi ,\csT) node[midway,font=\tiny]{$\mathcal{S}^{(k)}_{\mathrm{inacc}}$};
  \draw[partP ] (\cSi ,\csB) rectangle (\cPi,\csT) node[midway,font=\tiny]{instr};
  \draw[partPy] (\cPi,\csB) rectangle (\cPy,\csT) node[midway,font=\tiny,text=green!30!black]{$\mathbf{y}_{1:m_k}$};
  \draw[partD ] (\cPy,\csB) rectangle (\mR  ,\csT) node[midway,font=\tiny]{$\mathcal{D}$};
  \draw[black!50, line width=0.3pt] (\cSi, \csB-0.05) -- (\cSi, \csB-0.14) -- (\cPy, \csB-0.14) -- (\cPy, \csB-0.05);
  \node[labelnote, anchor=north, font=\tiny] at ({(\cSi+\cPy)/2}, \csB-0.15) {instr $+$ $\mathbf{y}_{1:m_k}$};
  \draw[captureline] ({\mL-0.05}, {\csT+0.14}) -- ({\mR+0.05}, {\csT+0.14});
  \draw[captureline] ({\mL-0.05}, {\csT+0.06}) -- ({\mL-0.05}, {\csT+0.14});
  \draw[captureline] ({\mR+0.05}, {\csT+0.06}) -- ({\mR+0.05}, {\csT+0.14});
  \node[labelnote, anchor=south] at ({(\mL+\mR)/2}, {\csT+0.18})
    {\textcolor{orange!70!black}{\textbf{capture all} $\mathbf{K}^{(\ell,h)}$ columns}};

  \def\rsL{1.20}\def\rsR{1.52}
  \draw[partP ] (\rsL,\rPa ) rectangle (\rsR,\mT  ) node[midway,rotate=90,font=\tiny]{$\mathcal{P}_1$};
  \draw[partSa] (\rsL,\rSa ) rectangle (\rsR,\rPa ) node[midway,rotate=90,font=\tiny]{$\mathcal{S}^{(k)}_{\mathrm{acc}}$};
  \draw[partSi] (\rsL,\rSi ) rectangle (\rsR,\rSa ) node[midway,rotate=90,font=\tiny]{$\mathcal{S}^{(k)}_{\mathrm{inacc}}$};
  \draw[partP ] (\rsL,\rPi) rectangle (\rsR,\rSi ) node[midway,rotate=90,font=\tiny]{instr};
  \draw[partPy] (\rsL,\rPy) rectangle (\rsR,\rPi) node[midway,rotate=90,font=\tiny,text=green!30!black]{$\mathbf{y}_{1:m_k}$};
  \draw[partD ] (\rsL,\mB  ) rectangle (\rsR,\rPy) node[midway,rotate=90,font=\tiny]{$\mathcal{D}$};
  \node[labelnote, rotate=90] at ({\rsL-0.58}, {(\mB+\mT)/2})
    {query rows in the full prompt};

  \draw[captureline] ({\rsL-0.14}, \mB) -- ({\rsL-0.14}, \rPy);
  \draw[captureline] ({\rsL-0.14}, \mB) -- ({\rsL-0.06}, \mB);
  \draw[captureline] ({\rsL-0.14}, \rPy) -- ({\rsL-0.06}, \rPy);
  \node[labelnote, text=orange!70!black, anchor=east, align=right, text width=1.55cm]
    at (0.30, {(\mB+\rPy)/2})
    {\textbf{capture only}\\ $\mathbf{Q}^{(\ell,h)}_{\mathcal{D}}$\\(draft rows)};

  \draw[panel] (10.3, 0.35) rectangle (15.9, 6.85);
  \node[title, anchor=north west] at (10.55, 6.75) {2.~Recompute Only That Block};
  \node[subtitle, anchor=north west, text width=5.05cm] at (10.55, 6.29)
    {Exact \emph{qk-fast} recomputation from captured $Q/K$; then average over $\mathcal{H}$, split accessible vs.\ inaccessible source mass, and apply the acceptance gates.};

  \def\bL{11.25}\def\bR{15.45}\def\bMid{13.48}
  \def\bB{0.95}\def\bT{3.15}
  \def\bcB{3.28}\def\bcT{3.60}

  \draw[partSa] (\bL,\bcB) rectangle (\bMid,\bcT);
  \node[labelnote] at ({(\bL+\bMid)/2}, {(\bcB+\bcT)/2}) {$\mathcal{S}^{(k)}_{\mathrm{acc}}$};
  \draw[partSi] (\bMid,\bcB) rectangle (\bR,\bcT);
  \node[labelnote] at ({(\bMid+\bR)/2}, {(\bcB+\bcT)/2}) {$\mathcal{S}^{(k)}_{\mathrm{inacc}}$};
  \node[labelnote, anchor=south] at ({(\bL+\bR)/2}, {\bcT+0.04})
    {source columns in local live-source order};

  \def\brL{10.92}\def\brR{11.18}
  \draw[partD] (\brL,\bB) rectangle (\brR,\bT);
  \node[labelnote, rotate=90] at ({(\brL+\brR)/2}, {(\bB+\bT)/2}) {$\mathcal{D}$};
  \node[labelnote, rotate=90] at (10.64, {(\bB+\bT)/2})
    {draft query rows};

  \draw[reconbox] (\bL,\bB) rectangle (\bR,\bT);
  \draw[teal!55!black, dashed, line width=0.4pt] (\bMid,\bB) -- (\bMid, 1.68);
  \draw[teal!55!black, dashed, line width=0.4pt] (\bMid, 2.42) -- (\bMid,\bT);
  \node[font=\small, text=teal!55!black]
    at ({(\bL+\bR)/2}, {(\bB+\bT)/2 + 0.18})
    {$\widehat{\mathbf{A}}^{(\ell,h)}_{\mathcal{D},\,\mathcal{S}^{(k)}}$};
  \node[labelnote, text=teal!55!black]
    at ({(\bL+\bR)/2}, {(\bB+\bT)/2 - 0.20})
    {$L_k$ draft rows $\times$ $N_k$ source columns};

  \draw[stagearrow]
    (9.35, {(\bB+\bT)/2})
    -- (11.08, {(\bB+\bT)/2});
\end{tikzpicture}
}%
\caption{\textbf{Selective reconstruction with runtime capture.} The
$n{\times}n$ attention matrix
$\mathbf{A}^{(\ell,h)}$ is executed entirely inside the fused attention
kernel, so the full $n{\times}n$ matrix is never materialized. For each
AlignAtt head $(\ell,h)\in\mathcal{H}$, the observer copies
$\mathbf{K}^{(\ell,h)}$ for all positions and $\mathbf{Q}^{(\ell,h)}$ only for
the draft rows $\mathcal{D}$ into fixed-shape buffers. The green band
$\mathbf{y}_{1:m_k}$ marks the committed target prefix that grows across
chunks; hatched cells are never reconstructed. After the forward,
Eq.~\eqref{eq:mt-qk-fast} recomputes exactly only the draft-to-source block
$\widehat{\mathbf{A}}_{\mathcal{D},\mathcal{S}^{(k)}}$, which is the sole
object consumed by the source-side provenance split of
Eqs.~\eqref{eq:mt-prov-acc} to~\eqref{eq:mt-provenance}, where
$\pi_t^{\mathrm{acc}}$ drives the gate and $\pi_t^{\mathrm{inacc}}$ is
retained for diagnostics, and by the acceptance policy of this section.}
\label{fig:qk-fast}
\end{figure*}
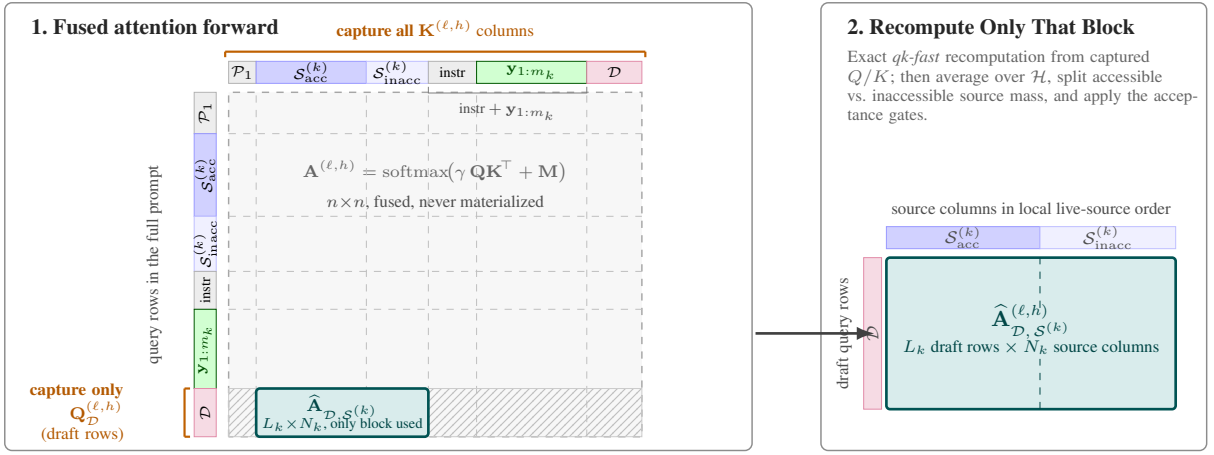

The policy never needs the full $n\times n$ self-attention matrix. It only
needs draft rows against source columns. For selected heads $(\ell,h)$, draft
positions $t$, and source positions $s$, we reconstruct
\begin{equation}
\widehat{A}^{(\ell,h)}_{t,s}
=
\frac{\exp\!\left(
\gamma^{(\ell)} q^{(\ell,h)}_t \cdot k^{(\ell,h)}_{\phi^{(k)}(s)}
  + m_{t,\phi^{(k)}(s)}
\right)}
{\sum_j \exp\!\left(
\gamma^{(\ell)} q^{(\ell,h)}_t \cdot k^{(\ell,h)}_j + m_{t,j}
\right)},
\label{eq:mt-qk-fast}
\end{equation}
where $q$ and $k$ are the captured post-normalization tensors after rotary
position embedding (RoPE; \citealp{su2021roformer}) actually consumed by the
deployed attention module, $\gamma^{(\ell)}$ is that
module's runtime scaling factor, and $m_{t,j}$ replays the same causal and
sliding-window masks as the fused forward. We call this replay path
\emph{qk-fast}: it reconstructs only the policy-visible block and matches the
deployed attention algebra up to floating-point reassociation relative to the
fused forward.
\Cref{fig:qk-fast} illustrates the capture-and-replay path: the fused runtime
keeps the full attention matrix internal, while the observer stores only the
selected draft queries and prompt keys needed to recompute the source-facing
policy block.

From these per-head rows we build two source-side provenance features for each
drafted token: accessible-source mass $\pi_t^{\mathrm{acc}}$ and
inaccessible-source mass $\pi_t^{\mathrm{inacc}}$,
\begin{align}
  \pi_t^{\mathrm{acc}}
  &=
  \sum_{s \in \mathcal{S}^{(k)}_{\mathrm{acc}}} \bar{\mathbf{r}}_t(s),
  \label{eq:mt-prov-acc}\\
  \pi_t^{\mathrm{inacc}}
  &=
  \sum_{s \in \mathcal{S}^{(k)}_{\mathrm{inacc}}} \bar{\mathbf{r}}_t(s),
  \label{eq:mt-provenance}
\end{align}
where $\bar{\mathbf{r}}_t$ is the head-averaged reconstructed row. We
additionally normalize per-head rows online with prefix Welford statistics and
apply a short median filter along the source axis before taking the source-side
argmax; these operations stabilize the alignment peak without changing the
underlying replayed rows. The runtime gate only thresholds
$\pi_t^{\mathrm{acc}}$; the complementary $\pi_t^{\mathrm{inacc}}$ is retained
as a diagnostic split in Appendix~\ref{sec:appendix-observer}. 

\subsection{AlignAtt Acceptance Policy}

The MT policy is a first-failure scan over drafted tokens. Let
$\hat{s}_t$ be the source-side argmax of the filtered row for draft token $t$,
and let $N^{(k)}_{\mathrm{acc}}$ be the number of source words whose ASR end
time is already on the accessible side of the frontier. The scan stops when one
of three conditions first fires:
\begin{align}
  \text{\textsc{source-frontier}} :\quad
  &\hat{s}_t \ge N^{(k)}_{\mathrm{acc}} + b,
  \label{eq:mt-stop-frontier}\\
  \text{\textsc{argmax-mass-weak}} :\quad
  &\bar{\mathbf{r}}_t(\hat{s}_t) < \tau_{\mathrm{argmax}},
  \label{eq:mt-stop-argmax}\\
  \text{\textsc{provenance-weak}} :\quad
  &\pi_t^{\mathrm{acc}} < \tau_{\mathrm{src}}.
  \label{eq:mt-stop-prov}
\end{align}
In the official 850/1500 ms operating points, $b=1$,
$\tau_{\mathrm{argmax}}=0$, and $\tau_{\mathrm{src}}=0$, so the optional
argmax-mass and minimum-source-mass gates are present in the runtime but
inactive in these operating points.

\begin{figure*}[t]
\centering
\resizebox{\linewidth}{!}{%
\begin{tikzpicture}[
  x=1cm, y=1cm,
  >={Latex[length=2.4mm]},
  every node/.style={font=\small, inner sep=0pt},
  panel/.style={draw=black!45, rounded corners=3pt, line width=0.6pt, fill=white},
  branchL/.style={draw=black!45, line width=0.6pt, rounded corners=2.5pt,
                  fill=black!2},
  branchR/.style={draw=black!45, line width=0.6pt, rounded corners=2.5pt,
                  fill=black!2},
  gatebox/.style={draw=black!50, line width=0.7pt, rounded corners=2.5pt,
                  fill=black!2},
  tensorbox/.style={draw=black!50, line width=0.7pt, rounded corners=2pt,
                    fill=black!2},
  flow/.style={->, draw=black!75, line width=0.9pt},
  sectitle/.style={font=\scriptsize\bfseries, align=left, inner sep=0pt},
  caption/.style={font=\scriptsize, align=left, text=black!60, inner sep=0pt},
  eqn/.style={font=\scriptsize, align=left, text=black!85, inner sep=0pt},
  sublab/.style={font=\tiny, align=left, text=black!55, inner sep=0pt},
  axlab/.style={font=\tiny, align=center, text=black!55, inner sep=0pt},
]

  \def\Wmax{17.00}\def\Hmax{5.25}
  \draw[panel] (0.00,0.00) rectangle (\Wmax,\Hmax);

  \def\ixL{0.55}\def\ixR{2.25}\def\iyB{1.30}\def\iyT{4.50}

  \node[sectitle, text=black!80, anchor=south]
    at ({(\ixL+\ixR)/2 + 0.35},\iyT+0.35)
    {Input: reconstructed block};

  \draw[tensorbox] (\ixL,\iyB) rectangle (\ixR,\iyT);
  \foreach \i in {1,...,8} {
    \pgfmathsetmacro\yy{\iyB + (\iyT-\iyB)*\i/9}
    \draw[black!18, line width=0.25pt] (\ixL,\yy) -- (\ixR,\yy);
  }
  \foreach \j in {1,...,6} {
    \pgfmathsetmacro\xx{\ixL + (\ixR-\ixL)*\j/7}
    \draw[black!18, line width=0.25pt] (\xx,\iyB) -- (\xx,\iyT);
  }
  \draw[black!70, dashed, line width=0.5pt]
    ({\ixL + (\ixR-\ixL)*4/7},\iyB-0.05)
    -- ({\ixL + (\ixR-\ixL)*4/7},\iyT+0.05);
  \node[axlab, anchor=south] at ({(\ixL+\ixR)/2}, \iyT+0.05) {source $s$};
  \node[axlab, rotate=90, anchor=south] at ({\ixL-0.22}, {(\iyB+\iyT)/2})
    {draft $t$};
  \node[axlab, text=blue!55!black, anchor=north]
    at ({\ixL + (\ixR-\ixL)*2/7}, \iyB-0.02) {acc};
  \node[axlab, text=blue!40!black, anchor=north]
    at ({\ixL + (\ixR-\ixL)*5.5/7}, \iyB-0.02) {inacc};
  \node[eqn, anchor=north] at ({(\ixL+\ixR)/2}, \iyB-0.45)
    {$\widehat{\mathbf{A}}^{(\ell,h)}_{t,s},\ (\ell,h)\in\mathcal{H}$};
  \node[sublab, anchor=north] at ({(\ixL+\ixR)/2}, \iyB-0.90)
    {Eq.~\eqref{eq:mt-qk-fast}};

  \def\jx{2.90}
  \draw[flow, -] (\ixR+0.05,{(\iyB+\iyT)/2}) -- (\jx,{(\iyB+\iyT)/2});
  \draw[black!75, line width=0.85pt, -] (\jx,{(\iyB+\iyT)/2}) -- (\jx,4.00);
  \draw[black!75, line width=0.85pt, -] (\jx,{(\iyB+\iyT)/2}) -- (\jx,1.50);
  \draw[flow] (\jx,4.00) -- (3.50,4.00);
  \draw[flow] (\jx,1.50) -- (3.50,1.50);

  \def\aL{3.55}\def\aR{10.70}\def\aB{3.00}\def\aT{5.00}
  \draw[branchL] (\aL,\aB) rectangle (\aR,\aT);

  \node[sectitle, text=black!85, anchor=north west] at (\aL+0.20,\aT-0.15)
    {A.~Accessible-source mass};
  \node[caption, text=black!60, anchor=north west,
        text width=6.75cm] at (\aL+0.20,\aT-0.50)
    {sum the head-averaged row on the accessible side of the frontier};

  \node[eqn, anchor=west] at (\aL+0.20,\aT-1.30)
    {$\displaystyle \pi^{\mathrm{acc}}_t\;=\!\!\sum_{s\,\in\,\mathcal{S}^{(k)}_{\mathrm{acc}}}\!\!p_{t,s},
      \qquad p_{t,s}=\tfrac{1}{|\mathcal{H}|}\!\!\sum_{(\ell,h)\in\mathcal{H}}\!\!\widehat{\mathbf{A}}^{(\ell,h)}_{t,s}$};

  \draw[flow] (\aR,{(\aB+\aT)/2}) -- (\aR+0.55,{(\aB+\aT)/2});

  \def\bL{3.55}\def\bR{10.70}\def\bB{0.40}\def\bT{2.60}
  \draw[branchR] (\bL,\bB) rectangle (\bR,\bT);

  \node[sectitle, text=black!85, anchor=north west] at (\bL+0.20,\bT-0.15)
    {B.~Stabilized source peak};
  \node[caption, text=black!60, anchor=north west,
        text width=6.75cm] at (\bL+0.20,\bT-0.50)
    {argmax on a normalized, smoothed row $\tilde z$ of the same heads};

  \node[eqn, anchor=west] at (\bL+0.20,\bT-1.30)
    {$\displaystyle \hat s_t\;=\;\arg\max_{s\,\in\,\mathcal{S}^{(k)}}\,\tilde z_{t,s}$};
  \node[sublab, anchor=west, text width=6.75cm] at (\bL+0.20,\bT-1.80)
    {$\tilde z$: per-head prefix-online z-score, averaged over $\mathcal{H}$,
     then width-$7$ median filter};

  \draw[flow] (\bR,{(\bB+\bT)/2}) -- (\bR+0.55,{(\bB+\bT)/2});

  \def\gL{11.30}\def\gR{14.35}\def\gB{0.40}\def\gT{5.00}
  \draw[gatebox] (\gL,\gB) rectangle (\gR,\gT);

  \node[sectitle, text=black!85, anchor=north west] at (\gL+0.20,\gT-0.15)
    {Gate $g_t$};
  \node[caption, text=black!60, anchor=north west,
        text width=2.65cm] at (\gL+0.20,\gT-0.50)
    {emit $y_t$ iff all three hold:};

  \node[eqn, anchor=west] at (\gL+0.20,\gT-1.20)
    {(i)\ \ $\hat s_t \,<\, N^{(k)}_{\mathrm{acc}} + b$};
  \node[sublab, anchor=west, text width=2.50cm]
    at (\gL+0.40,\gT-1.60) {peak stays left of the frontier};

  \node[eqn, anchor=west] at (\gL+0.20,\gT-2.30)
    {(ii)\ \ $p_{t,\hat s_t}\,\ge\,\tau_{\mathrm{argmax}}$};
  \node[sublab, anchor=west, text width=2.50cm]
    at (\gL+0.40,\gT-2.70) {local peak mass high enough};

  \node[eqn, anchor=west] at (\gL+0.20,\gT-3.40)
    {(iii)\ \ $\pi^{\mathrm{acc}}_t\,\ge\,\tau_{\mathrm{src}}$};
  \node[sublab, anchor=west, text width=2.50cm]
    at (\gL+0.40,\gT-3.80) {enough accessible mass overall};

  \node[sublab, text=black!50, anchor=south, text width=2.65cm,
        align=center] at ({(\gL+\gR)/2},\gB+0.10)
    {optional thresholds; inactive in official runs};

  \draw[flow] (\gR,{(\gB+\gT)/2}) -- (\gR+0.55,{(\gB+\gT)/2});

  \def\sL{14.95}\def\sR{16.85}
  \node[sectitle, anchor=north west] at (\sL,4.60) {Scan};
  \node[sublab, anchor=north west, text width=1.85cm] at (\sL,4.25)
    {left$\to$right; emit while $g_t{=}1$, drop at first $g_t{=}0$};

  \def\cellw{0.28}\def\cellh{0.38}\def\cellg{0.02}
  \def\stX{\sL}
  \def\stY{2.51}
  \foreach \i in {0,1,2} {
    \pgfmathsetmacro\x{\stX + (\cellw+\cellg)*\i}
    \filldraw[fill=green!22, draw=green!45!black, line width=0.4pt]
      (\x,\stY) rectangle (\x+\cellw,\stY+\cellh);
    \node[font=\tiny, text=green!30!black, inner sep=0pt]
      at (\x+\cellw/2,\stY+\cellh/2) {\textbf{\checkmark}};
  }
  \pgfmathsetmacro\xf{\stX + (\cellw+\cellg)*3}
  \filldraw[fill=red!18, draw=red!65!black, line width=0.55pt]
    (\xf,\stY) rectangle (\xf+\cellw,\stY+\cellh);
  \node[font=\tiny, text=red!60!black, inner sep=0pt]
    at (\xf+\cellw/2,\stY+\cellh/2) {\textbf{$\times$}};
  \foreach \i in {4,5} {
    \pgfmathsetmacro\x{\stX + (\cellw+\cellg)*\i}
    \filldraw[fill=black!5, draw=black!30, line width=0.35pt]
      (\x,\stY) rectangle (\x+\cellw,\stY+\cellh);
    \node[font=\tiny, text=black!45, inner sep=0pt]
      at (\x+\cellw/2,\stY+\cellh/2) {$\cdot$};
  }
  \foreach \i/\tok in {0/{$y_1$},1/{$y_2$},2/{$y_3$},3/{$y_4$},4/{$y_5$},5/{$y_6$}} {
    \pgfmathsetmacro\x{\stX + (\cellw+\cellg)*\i}
    \node[font=\tiny, text=black!70, inner sep=0pt]
      at (\x+\cellw/2,\stY-0.22) {\tok};
  }
  \node[sublab, text=green!30!black, anchor=north west, text width=1.1cm]
    at (\sL,\stY-0.50) {emit $y_1{:}y_3$};
  \node[sublab, text=red!60!black, anchor=north west, text width=0.9cm]
    at ({\sL+1.10},\stY-0.50) {drop rest};

\end{tikzpicture}
}%
\caption{\textbf{From the reconstructed block to an acceptance decision.}
The selective \emph{qk-fast} reconstruction produces
$\widehat{\mathbf{A}}^{(\ell,h)}_{t,s}$ for $(\ell,h)\in\mathcal{H}$, with draft
positions $t$ against source positions $s$ (left). The policy reads it through
two parallel aggregations. \emph{Branch~A} averages the selected heads into a
row $p_{t,\cdot}$ and sums its mass on the accessible side of the frontier,
giving the provenance score $\pi^{\mathrm{acc}}_t$. \emph{Branch~B} returns
the peak $\hat s_t$ of a stabilized row $\tilde z_{t,\cdot}$, where $\tilde
z$ is obtained by z-scoring each head with prefix-online Welford moments,
averaging over~$\mathcal{H}$, and applying a width-$7$ median filter along
the source axis. The gate $g_t$ accepts draft token $t$ iff the peak stays on
the accessible side of the frontier, the peak mass exceeds
$\tau_{\mathrm{argmax}}$, and the accessible-source mass exceeds
$\tau_{\mathrm{src}}$. A left-to-right scan emits the longest run of accepting
tokens and drops the remainder at the first failure.}
\label{fig:policy-pipeline}
\end{figure*}
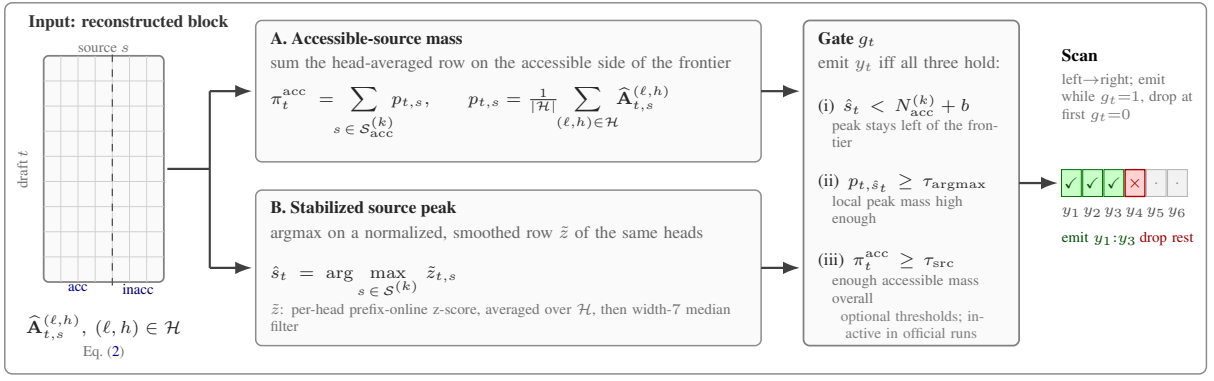

Figure~\ref{fig:policy-pipeline} makes the decision path explicit. One branch
averages the retained heads and sums accessible-source mass to obtain
$\pi_t^{\mathrm{acc}}$. The other normalizes and smooths the same replayed rows
before taking the source-side argmax $\hat{s}_t$. The gate accepts token $t$
if the three tests in
Eqs.~\eqref{eq:mt-stop-frontier} to~\eqref{eq:mt-stop-prov} all pass; a
left-to-right first-failure scan then emits the longest accepting draft prefix
and rounds back to the last completed stability unit. Here a stability unit is
the smallest target fragment that the tokenizer treats as safe to commit: a
whitespace-delimited lexical word in spacing scripts such as EN$\to$DE/IT, or
a single CJK character in EN$\to$ZH, so partial subword fragments are never
emitted.

\subsection{Runtime Query/Key Capture}

Equation~\eqref{eq:mt-qk-fast} is only useful if the deployed runtime exposes
the exact queries and keys consumed by attention. In the deployed vLLM path,
those tensors never appear as ordinary Python-visible objects. We therefore
install a per-layer observer, copy prompt keys and draft-row queries into
fixed-shape buffers, and route capture through a custom-op-style call that
survives runtime graph lowering. A zero-valued sentinel
dependency keeps the observer in the transitive fan-in of the graph output:
\begin{equation}
  \mathbf{o}'_\ell =
  \mathbf{o}_\ell +
  \Phi\bigl(\ell,\mathrm{pos},\mathbf{Q}^{(\ell)},\mathbf{K}^{(\ell)}\bigr),
  \qquad \Phi \equiv \mathbf{0}.
  \label{eq:mt-sentinel}
\end{equation}

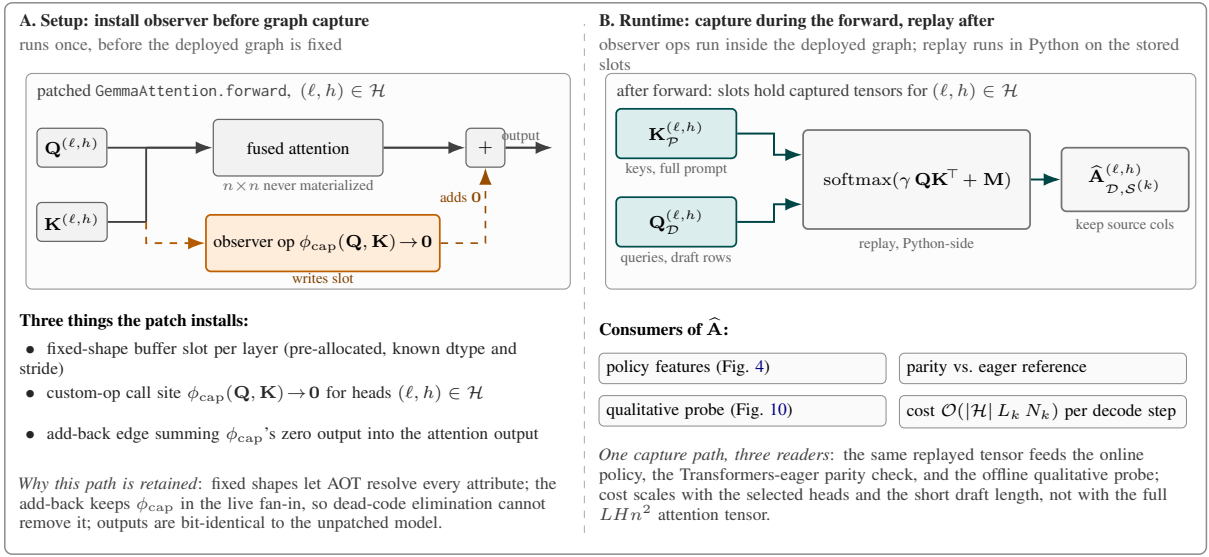
\begin{figure*}[t]
\centering
\resizebox{\linewidth}{!}{%
\begin{tikzpicture}[
  x=1cm, y=1cm,
  >={Latex[length=2.4mm]},
  every node/.style={font=\small, inner sep=0pt},
  panel/.style={draw=black!45, rounded corners=3pt, line width=0.6pt, fill=white},
  inner/.style={draw=black!40, line width=0.45pt, rounded corners=2.5pt,
                fill=black!1},
  opnode/.style={draw=black!60, fill=black!5, rounded corners=2pt,
                 line width=0.5pt, inner sep=3pt, font=\scriptsize,
                 align=center},
  obsnode/.style={draw=orange!75!black, fill=orange!14, rounded corners=2pt,
                  line width=0.7pt, inner sep=3pt, font=\scriptsize,
                  align=center},
  bufnode/.style={draw=teal!60!black, fill=teal!14, rounded corners=2pt,
                  line width=0.7pt, inner sep=3pt, font=\scriptsize,
                  align=center},
  outnode/.style={draw=black!55, fill=black!3, rounded corners=2pt,
                  line width=0.7pt, inner sep=3pt, font=\scriptsize,
                  align=center},
  tagbox/.style={draw=black!45, fill=black!2, rounded corners=2pt,
                 line width=0.4pt, inner sep=3pt, font=\scriptsize, align=left},
  tflow/.style={->, draw=black!75, line width=0.9pt},
  dflow/.style={->, draw=teal!55!black, line width=0.8pt},
  ocap/.style={->, draw=orange!70!black, dashed, line width=0.75pt},
  ozero/.style={->, draw=orange!60!black, dashed, line width=0.65pt},
  wireblk/.style={draw=black!75, line width=0.75pt},
  sectitle/.style={font=\scriptsize\bfseries, align=left, inner sep=0pt},
  sub/.style={font=\scriptsize, align=left, text=black!60, inner sep=0pt},
  tny/.style={font=\tiny, align=left, text=black!55, inner sep=0pt},
  cen/.style={font=\tiny, align=center, text=black!60, inner sep=0pt},
  bullet/.style={font=\scriptsize, align=left, text=black!85, inner sep=0pt},
  prose/.style={font=\scriptsize, align=left, text=black!75, inner sep=0pt},
]

  \def\Wmax{17.00}\def\Hmax{7.80}
  \draw[panel] (0.00,0.00) rectangle (\Wmax,\Hmax);

  \draw[black!30, dashed, line width=0.45pt] (8.20,0.30) -- (8.20,7.50);

  \node[sectitle, text=black!85, anchor=north west] at (0.20,7.65)
    {A.~Setup: install observer before graph capture};
  \node[sub, anchor=north west, text width=7.80cm] at (0.20,7.30)
    {runs once, before the deployed graph is fixed};

  \def\xLL{0.30}\def\xLR{8.00}\def\xLB{3.75}\def\xLT{6.80}
  \draw[inner] (\xLL,\xLB) rectangle (\xLR,\xLT);
  \node[sub, text=black!75, anchor=north west] at (\xLL+0.15,\xLT-0.12)
    {patched \texttt{GemmaAttention.forward},\ \ $(\ell,h)\in\mathcal{H}$};

  \node[opnode, minimum width=0.85cm, minimum height=0.55cm]
    (Q) at (0.95,5.75) {$\mathbf{Q}^{(\ell,h)}$};
  \node[opnode, minimum width=0.85cm, minimum height=0.55cm]
    (K) at (0.95,4.70) {$\mathbf{K}^{(\ell,h)}$};

  \def\Bx{2.00}
  \draw[wireblk] (Q.east) -- (\Bx,5.75);
  \draw[wireblk] (K.east) -- (\Bx,4.70);
  \draw[wireblk] (\Bx,5.75) -- (\Bx,4.70);

  \node[opnode, minimum width=2.4cm, minimum height=0.80cm]
    (fused) at (4.15,5.75) {fused attention};
  \node[cen, anchor=north] at (4.15,5.30)
    {$n{\times}n$ never materialized};

  \node[obsnode, minimum width=3.15cm, minimum height=0.80cm, align=center]
    (obs) at (4.50,4.40)
    {observer op $\phi_{\mathrm{cap}}(\mathbf{Q},\mathbf{K})\!\to\!\mathbf{0}$};
  \node[cen, anchor=north, text=orange!50!black] at (4.50,3.98)
    {writes slot};

  \node[opnode, minimum width=0.55cm, minimum height=0.55cm]
    (sum) at (6.80,5.75) {$+$};

  \draw[tflow] (\Bx,5.75) -- (fused.west);
  \draw[ocap] (\Bx,4.70) -- (\Bx,4.40) -- (obs.west);

  \draw[tflow] (fused.east) -- (sum.west);

  \draw[ozero] (obs.east) -- (6.80,4.40) -- (sum.south);
  \node[cen, text=orange!55!black, anchor=east] at (6.75,5.05)
    {adds $\mathbf{0}$};

  \draw[tflow] (sum.east) -- (7.75,5.75);
  \node[cen, anchor=south] at (7.30,5.80) {output};

  \node[sectitle, anchor=north west] at (0.20,3.40)
    {Three things the patch installs:};

  \node[bullet, anchor=north west, text width=7.60cm] at (0.20,3.00)
    {\ $\bullet$\ \ fixed-shape buffer slot per layer (pre-allocated, known
     dtype and stride)};
  \node[bullet, anchor=north west, text width=7.60cm] at (0.20,2.40)
    {\ $\bullet$\ \ custom-op call site
     $\phi_{\mathrm{cap}}(\mathbf{Q},\mathbf{K})\!\to\!\mathbf{0}$ for heads
     $(\ell,h)\in\mathcal{H}$};
  \node[bullet, anchor=north west, text width=7.60cm] at (0.20,1.80)
    {\ $\bullet$\ \ add-back edge summing $\phi_{\mathrm{cap}}$'s zero output
     into the attention output};

  \node[prose, anchor=north west, text width=7.60cm] at (0.20,1.10)
    {\emph{Why this path is retained}: fixed shapes let AOT resolve every
     attribute; the add-back keeps $\phi_{\mathrm{cap}}$ in the live fan-in,
     so dead-code elimination cannot remove it; outputs are bit-identical to
     the unpatched model.};

  \node[sectitle, text=black!85, anchor=north west] at (8.40,7.65)
    {B.~Runtime: capture during the forward, replay after};
  \node[sub, anchor=north west, text width=8.40cm] at (8.40,7.30)
    {observer ops run inside the deployed graph; replay runs in Python on the
     stored slots};

  \def\xRL{8.50}\def\xRR{16.85}\def\xRB{3.75}\def\xRT{6.80}
  \draw[inner] (\xRL,\xRB) rectangle (\xRR,\xRT);
  \node[sub, text=black!75, anchor=north west] at (\xRL+0.15,\xRT-0.12)
    {after forward: slots hold captured tensors for $(\ell,h)\in\mathcal{H}$};

  \node[bufnode, minimum width=1.70cm, minimum height=0.70cm]
    (bK) at (9.50,5.95) {$\mathbf{K}^{(\ell,h)}_{\mathcal{P}}$};
  \node[cen, anchor=north] at (9.50,5.54) {keys, full prompt};

  \node[bufnode, minimum width=1.70cm, minimum height=0.70cm]
    (bQ) at (9.50,4.70) {$\mathbf{Q}^{(\ell,h)}_{\mathcal{D}}$};
  \node[cen, anchor=north] at (9.50,4.29) {queries, draft rows};

  \node[outnode, text width=3.00cm, align=center, minimum height=1.40cm]
    (replay) at (12.90,5.30)
    {$\mathrm{softmax}(\gamma\,\mathbf{Q}\mathbf{K}^{\!\top}\!+\mathbf{M})$};
  \node[cen, anchor=north] at (12.90,4.48) {replay, Python-side};

  \node[outnode, minimum width=1.80cm, minimum height=0.90cm]
    (slice) at (15.85,5.30)
    {$\widehat{\mathbf{A}}^{(\ell,h)}_{\mathcal{D},\mathcal{S}^{(k)}}$};
  \node[cen, anchor=north] at (15.85,4.75) {keep source cols};

  \draw[dflow] (bK.east) -| ([xshift=-0.40cm, yshift=0.35cm]replay.west)
               -- ([yshift=0.35cm]replay.west);
  \draw[dflow] (bQ.east) -| ([xshift=-0.40cm, yshift=-0.35cm]replay.west)
               -- ([yshift=-0.35cm]replay.west);
  \draw[dflow] (replay.east) -- (slice.west);

  \node[sectitle, anchor=north west] at (8.40,3.35)
    {Consumers of $\widehat{\mathbf{A}}$:};

  \node[tagbox, anchor=north west, text width=3.85cm] at (8.40,2.85)
    {policy features (Fig.~\ref{fig:policy-pipeline})};
  \node[tagbox, anchor=north west, text width=3.85cm] at (12.65,2.85)
    {parity vs.\ eager reference};
  \node[tagbox, anchor=north west, text width=3.85cm] at (8.40,2.25)
    {qualitative probe (Fig.~\ref{fig:mt-selective-reconstruction})};
  \node[tagbox, anchor=north west, text width=3.85cm] at (12.65,2.25)
    {cost $\mathcal{O}(|\mathcal{H}|\,L_k\,N_k)$ per decode step};

  \node[prose, anchor=north west, text width=8.20cm] at (8.40,1.50)
    {\emph{One capture path, three readers}: the same replayed tensor feeds
     the online policy, the Transformers-eager parity check, and the offline
     qualitative probe; cost scales with the selected heads and the short
     draft length, not with the full $LHn^2$ attention tensor.};

\end{tikzpicture}
}%
\caption{\textbf{Observer lifecycle on the deployed vLLM path.}
\emph{Left (A), once, before graph capture:} the forward of
\texttt{GemmaAttention} is patched so that for each selected head
$(\ell,h)\in\mathcal{H}$ the query and key tensors also pass through an
observer custom op $\phi_{\mathrm{cap}}$ (dashed orange side path). The op
writes the selected slices into fixed-shape, pre-allocated slots and returns
a zero tensor that is added back into the attention output; this keeps the op
in the live fan-in of the graph so graph optimization cannot dead-code-eliminate
it, while the numerical output of the layer is unchanged.
\emph{Right (B), every forward:} after the forward finishes, the
observer slots hold $\mathbf{K}^{(\ell,h)}_{\mathcal{P}}$ for the full prompt
and $\mathbf{Q}^{(\ell,h)}_{\mathcal{D}}$ for the draft rows; a short
Python-side replay reapplies the module scaling and the same prompt/suffix
masks used by the deployed attention kernel, then keeps the source columns to
yield the selective block
$\widehat{\mathbf{A}}^{(\ell,h)}_{\mathcal{D},\mathcal{S}^{(k)}}$ that feeds
the policy, the parity check against a Transformers reference,
and the qualitative probe.}
\label{fig:observer-lifecycle}
\end{figure*}

Figure~\ref{fig:observer-lifecycle} separates the capture path into three
stages. During setup, we patch the attention forward and install fixed-shape
observer slots before the graph is captured. During each forward, the live
observer records prompt keys and draft-row queries for the selected heads after
the module's $q/k$ normalization and rotary embedding, then returns
$\Phi \equiv \mathbf{0}$ back into the attention output. After the forward, a
short Python-side replay reconstructs only the selective
draft-to-source block consumed by the policy and by the parity suite. The
sentinel add-back of Eq.~\eqref{eq:mt-sentinel} is therefore only a liveness
device: without the additive zero, graph lowering dead-code-eliminates the
observer path and no usable Q/K buffers survive on the deployed runtime.
Appendix~\ref{sec:appendix-observer} keeps the replay equations, parity
measurements, and qualitative probe.

\section{Results}
\label{sec:results}

\subsection{Experimental Setup}
\label{sec:setup}

The experiments run on a single NVIDIA A40. The MT component uses Gemma-4 E4B-it
in bfloat16 through vLLM; the speech component uses Qwen3-ASR with
forced alignment. In evaluation, we use the official IWSLT 2026 MCIF dev set
($\sim 2.1$ hours total, 21 long academic talks) resegmented by OmniSTEval.
We report BLEU and chrF against the dev references, XCOMET-XL for
translation adequacy \citep{xcometxl24}, and LongYAAL latency in its CU and
CA variants \citep{longyaal25}. The streaming loop follows the
Simulstream setting of unsegmented long-form audio with revision-capable
evaluation \citep{simulstream25}, but our cascade itself is
synchronous: ASR and MT are serialized on one GPU so that policy effects are
not confounded with scheduler overlap.

Section~\ref{sec:runtime} compares three MT implementations on 16 text-only
prompts: Transformers eager, Transformers SDPA \emph{qk-fast}, and deployed
vLLM \emph{qk-fast}. All use greedy decoding with the same retained head set.

\subsection{System Results}
\label{sec:main-results}

Table~\ref{tab:e2e} reports the two official operating points together with the
organizers' no-context baselines for comparison.
The pattern is
consistent across EN$\to$DE, EN$\to$ZH, and EN$\to$IT: the low-latency regime
stays near a 2\,s CU-LongYAAL, while the high-latency regime moves to a
clearly stronger quality point at the expected latency cost. CA-LongYAAL is
consistently below CU-LongYAAL here because OmniSTEval's CA mode
replaces each chunk-boundary audio increment by the actual wall-clock increment
spent processing that chunk. Since the deployed system runs faster than real
time, the CA timestamps can be earlier than the CU chunk-boundary timestamps.
We read Table~\ref{tab:e2e} as evidence that the \AAName{}
realization supports
both a stable low-latency operating point and a higher-quality regime on the
same decoder-only Gemma backend.

\begin{table}[!t]
\centering
\footnotesize
\setlength{\tabcolsep}{2.5pt}
{%
\begin{tabular}{l l l c c c r r}
\toprule
Lang. & System & Reg. & BLEU & chrF & XCOM. & \multicolumn{2}{c}{L. YAAL} \\
\cmidrule(lr){7-8}
& & & & & & CU & CA \\
\midrule
\multirow{5}{*}{en$\to$de}
& Baseline & low  & $22.35$ & $56.7$ & $0.748$ & \textbf{1.81} & n/a \\
& Ours          & low  & \textbf{28.76} & \textbf{62.1} & \textbf{0.875} & $2.00$ & $1.63$ \\
& Baseline & high & $26.31$ & $59.2$ & $0.819$ & \textbf{2.63} & n/a \\
& Ours          & high & \textbf{32.63} & \textbf{64.2} & \textbf{0.902} & $3.53$ & $3.14$ \\
& \offlinecell{Offline} & \offlinecell{-} & \offlinecell{38.57} & \offlinecell{67.1} & \offlinecell{0.938} & \offlinecell{n/a} & \offlinecell{n/a} \\
\midrule
\multirow{5}{*}{en$\to$zh}
& Baseline & low  & \textbf{40.85} & $34.1$ & \textbf{0.750} & \textbf{1.91} & n/a \\
& Ours          & low  & $36.01$ & \textbf{35.0} & $0.743$ & $1.95$ & $1.77$ \\
& Baseline & high & \textbf{43.85} & \textbf{37.8} & \textbf{0.795} & $3.48$ & n/a \\
& Ours          & high & $39.86$ & \textbf{37.8} & $0.778$ & \textbf{3.27} & $3.09$ \\
& \offlinecell{Offline} & \offlinecell{-} & \offlinecell{48.53} & \offlinecell{43.4} & \offlinecell{0.848} & \offlinecell{n/a} & \offlinecell{n/a} \\
\midrule
\multirow{5}{*}{en$\to$it}
& Baseline & low  & $30.63$ & $62.0$ & $0.683$ & \textbf{1.76} & n/a \\
& Ours          & low  & \textbf{40.10} & \textbf{68.0} & \textbf{0.805} & $1.98$ & $1.62$ \\
& Baseline & high & $37.28$ & $65.4$ & $0.781$ & \textbf{3.30} & n/a \\
& Ours          & high & \textbf{44.46} & \textbf{70.1} & \textbf{0.841} & $3.48$ & $3.10$ \\
& \offlinecell{Offline} & \offlinecell{-} & \offlinecell{49.88} & \offlinecell{73.0} & \offlinecell{0.895} & \offlinecell{n/a} & \offlinecell{n/a} \\
\bottomrule
\end{tabular}
}
\caption{\textbf{Cascade results on the IWSLT 2026 dev set}
(21 clips, OmniSTEval long-form resegmentation). Ours low rows use the
low-latency setting with $\Delta_{\mathrm{chunk}}=850$ ms; Ours high rows
use $\Delta_{\mathrm{chunk}}=1500$ ms. Organizer baseline rows are
the no-context baseline outputs provided with CU-LongYAAL; their
CA-LongYAAL was not available. Offline rows are diagnostic
quality-only cascades with Qwen full-audio ASR followed by Gemma final-mode MT;
they have no streaming latency score.}
\label{tab:e2e}
\end{table}

Against the supplied organizers' no-context baselines, \AAName{} clearly wins
the $<2$\,s and $<4$\,s latency regimes for EN$\to$DE and EN$\to$IT.
EN$\to$ZH is mixed: our system reaches comparable chrF at high latency and
slightly higher chrF at low latency, but the organizers' baseline remains ahead
on BLEU and XCOMET-XL.
The offline diagnostic rows show substantial backbone headroom once online
commitment is removed. They are not latency-comparable to the streaming runs,
but they help separate model capacity from the cost of the
simultaneous policy. \Aref{sec:appendix-offline} gives the minimal setup
details. Gemma-4 is not necessarily the optimal MT backbone for this policy: recent translation-focused decoder-only LLMs such as HY-MT-1.5 \citep{hymt1526} and MiLMMT-46 \citep{milmmt4626} achieve strong multilingual results, especially for Chinese, and require no changes to the AlignAtt implementation, though the alignment head set and acceptance thresholds would need to be recalibrated for each new backbone.

\subsection{Additional Analysis: ASR Tail Reliability}
\label{sec:asr-tail-analysis}

The official runs expose each Qwen3 ASR word to MT as soon as its aligned
end time is observed. A post-hoc reliability analysis suggests a slightly more
conservative default for future runs: clip the last 250\,ms of the live ASR
tail before passing the source prefix to MT. This is a policy-level timing
margin rather than a lexical repair. Because the Qwen path re-transcribes the
live utterance tail from scratch at each chunk, the final ASR words are exactly
the part most likely to be revised on the next chunk. This instability is partly
masked by AlignAtt, which often stops MT before translating the newest source
words near the frontier. Reducing tail noise could therefore allow a less
conservative acceptance policy, not only a longer hold-back. On the 21 MCIF dev talks,
the reference error rate of words at the current Qwen3 ASR tail end is 17.1\%,
but drops to 8.3\% at 250\,ms and then remains nearly flat. We therefore keep
the reported numbers unchanged, but recommend a 250\,ms ASR-tail hold-back as
the maintained default going forward. \Aref{app:appendix-asr} gives
the comparison with the alternative ASR component parameters.

\begin{figure}[t]
\centering
\includegraphics[width=\columnwidth]{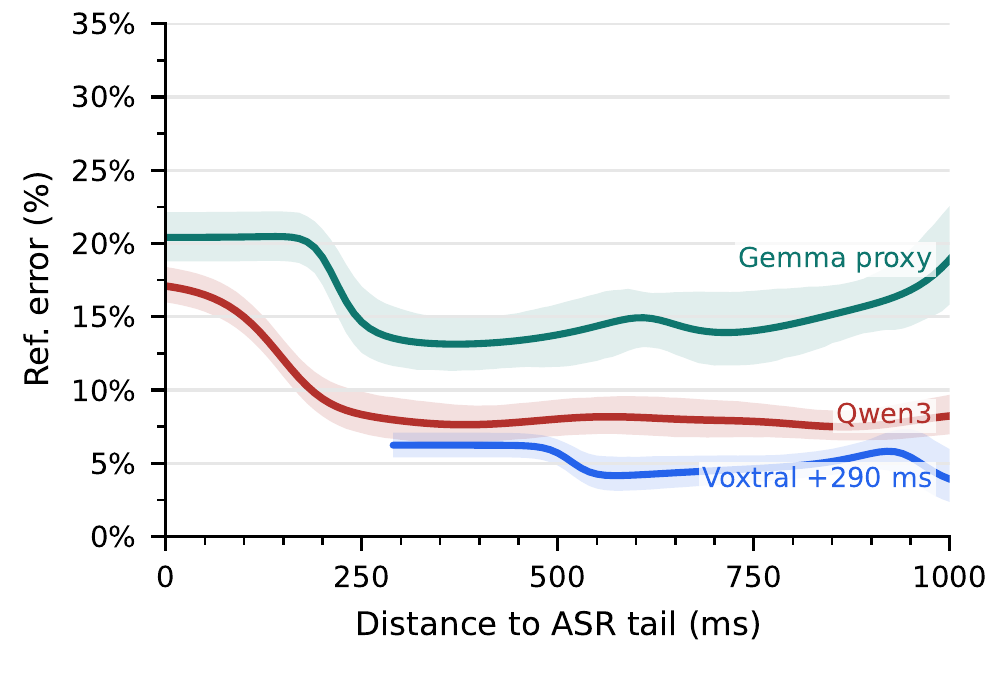}
\vspace{-1.0em}
\caption{\textbf{Live-tail ASR reference error.}
Reference-error rate by distance to the current ASR tail; bands are 90\%
audio-bootstrap intervals. Qwen3 drops from 17.1\% at the tail to 8.3\% at
250\,ms and then stays flat. Voxtral is shifted by +290\,ms CU-LongYAAL;
Gemma remains unshifted because its timestamps/LongYAAL are unreliable under
prompt leakage.}
\label{fig:asr-tail-risk}
\end{figure}

\subsection{Head Filtering Matters End-to-End}
\label{sec:head-filtering}

Table~\ref{tab:mt-e2e-head-filtering} is an auxiliary EN$\to$DE rerun included
only to isolate the MT head-set effect under the maintained \emph{qk-fast}
runtime. End-to-end quality is nearly equivalent, but the latency split is
informative. CU-LongYAAL increases by 100.3\,ms, which likely means the all-head
observer is slightly less precise and therefore a bit more generous on tokens
whose support extends beyond the currently available source prefix. CA-LongYAAL
rises more strongly, by +179.5\,ms, because replaying all 336 heads forces the
runtime to reconstruct much more attention state at each MT step. We therefore
read the auxiliary comparison as evidence that a small retained head set
captures most of the useful MT alignment signal at much lower runtime cost.

\begin{table}[t]
\centering
\footnotesize
\setlength{\tabcolsep}{4pt}
\begin{tabular}{l c c c}
\toprule
Setting & XCOMET-XL $\uparrow$ & CU (s) $\downarrow$ & CA (s) $\downarrow$ \\
\midrule
Top-8 heads & 0.879 & \textbf{1.96} & \textbf{1.65} \\
All 336 heads & \textbf{0.885} & 2.06 & 1.83 \\
\bottomrule
\end{tabular}
\caption{\textbf{Auxiliary EN$\to$DE MT head-set comparison at $\Delta_{\mathrm{chunk}}=1100$ ms.} Both rows use the same maintained runtime; only the MT head set changes. End-to-end quality is nearly equivalent, but all-head replay increases CU and especially CA because the observer becomes less selective while the runtime must reconstruct much more attention at each MT step.}
\label{tab:mt-e2e-head-filtering}
\end{table}

\subsection{Runtime Cost of the MT Component}
\label{sec:runtime}

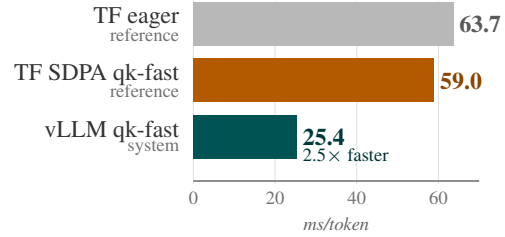
\begin{figure}[t]
\centering
\begin{tikzpicture}[
  x=1cm, y=1cm,
  >={Latex[length=2.2mm]},
  every node/.style={font=\footnotesize, inner sep=0pt},
  title/.style={font=\footnotesize\bfseries, text=black!85, align=left, inner sep=0pt},
  sub/.style={font=\scriptsize, text=black!60, align=left, inner sep=0pt},
  tck/.style={font=\scriptsize, text=black!55, align=center, inner sep=0pt},
  rowlbl/.style={font=\footnotesize, text=black!80, align=right, anchor=east, inner sep=0pt},
  rowsub/.style={font=\scriptsize, text=black!55, align=right, anchor=east, inner sep=0pt},
  valuelbl/.style={font=\footnotesize\bfseries, anchor=west, inner sep=0pt},
  grid/.style={draw=black!12, line width=0.35pt},
  axis/.style={draw=black!60, line width=0.55pt},
  barbase/.style={draw=none, line width=0pt},
]

  \def\Wmax{7.08}\def\Hmax{4.25}
  \path[use as bounding box] (0,0) rectangle (\Wmax,\Hmax);

  \node[title, anchor=north west] at (0.00,\Hmax-0.05)
    {Median MT decode latency};
  \node[sub, anchor=north west, text width=6.80cm] at (0.00,\Hmax-0.42)
    {Per generated token on 16 fixed text-only prompts.};

  \def\pxL{2.45}    
  \def\pxR{6.23}    
  \def\pyB{0.72}    
  \def\pyT{3.10}    
  \def\xScale{0.054}   

  \foreach \v in {0,20,40,60} {
    \pgfmathsetmacro\xv{\pxL + \xScale*\v}
    \draw[grid] (\xv,\pyB-0.04) -- (\xv,\pyT);
    \node[tck, anchor=north] at (\xv,\pyB-0.13) {\v};
  }
  \node[tck, anchor=north, font=\scriptsize\itshape, text=black!60]
    at ({(\pxL+\pxR)/2},\pyB-0.47) {ms/token};

  \draw[axis] (\pxL,\pyB) -- (\pxR,\pyB);

  \def\ryA{2.80}   
  \def\ryB{2.05}   
  \def\ryC{1.30}   
  \def\bh{0.29}    

  \node[rowlbl] at (\pxL-0.18,\ryA+0.08) {TF eager};
  \node[rowsub] at (\pxL-0.18,\ryA-0.14) {reference};
  \node[rowlbl] at (\pxL-0.18,\ryB+0.08) {TF SDPA qk-fast};
  \node[rowsub] at (\pxL-0.18,\ryB-0.14) {reference};
  \node[rowlbl] at (\pxL-0.18,\ryC+0.08) {vLLM qk-fast};
  \node[rowsub] at (\pxL-0.18,\ryC-0.14) {system};

  \pgfmathsetmacro\xA{\pxL + \xScale*63.7}
  \filldraw[barbase, fill=black!28] (\pxL,\ryA-\bh) rectangle (\xA,\ryA+\bh);
  \node[valuelbl, text=black!65] at (\xA+0.07,\ryA) {63.7};

  \pgfmathsetmacro\xB{\pxL + \xScale*59.0}
  \filldraw[barbase, fill=orange!70!black] (\pxL,\ryB-\bh) rectangle (\xB,\ryB+\bh);
  \node[valuelbl, text=orange!55!black] at (\xB+0.07,\ryB) {59.0};

  \pgfmathsetmacro\xC{\pxL + \xScale*25.4}
  \filldraw[barbase, fill=teal!65!black] (\pxL,\ryC-\bh) rectangle (\xC,\ryC+\bh);
  \node[valuelbl, text=teal!45!black] at (\xC+0.07,\ryC) {25.4};
  \node[font=\scriptsize, text=teal!40!black, anchor=west]
    at (\xC+0.07,\ryC-0.25) {2.5$\times$ faster};

\end{tikzpicture}
\caption{\textbf{Inference-time comparison of MT capture implementations.} Median
latency per generated token on a fixed 16-prompt text-only suite, for a
minimal Transformers eager reference, a Transformers SDPA
\emph{qk-fast} reference that reconstructs source rows from captured
layer inputs, and the deployed vLLM \emph{qk-fast} path used by the
presented system.}
\label{fig:mt-capture-speed}
\end{figure}

The MT component must be computationally cheap enough to justify deployment.
Figure~\ref{fig:mt-capture-speed} shows that the vLLM replay path is
much cheaper than the minimal eager implementation we use as an inspection reference:
median cost drops from 63.7 to 25.4 ms/token, while the Transformers
SDPA \emph{qk-fast} replay remains close to the eager baseline at
59.0 ms/token.
This confirms that the MT contribution is not merely inspectable in principle;
it is practical inside the runtime we actually deploy.

\section{Conclusion}
\label{sec:conclusion}

We presented a Qwen3 ASR + Gemma-4 MT cascade for simultaneous speech translation, and
its key technical ingredient is an AlignAtt policy adapted to decoder-only MT.
Deterministic prompt layout, offline head selection, selective attention
replay, and runtime query/key capture make the MT policy usable on the deployed
vLLM path. On the
dev set, this yields a low-latency operating point near 2\,s CU-LongYAAL and a
high-latency regime that is roughly 1.5\,s slower but clearly stronger in
BLEU and XCOMET-XL. Compared with the supplied organizers' no-context
baselines, the system clearly wins the $<2$\,s and $<4$\,s latency regimes for English to German and Italian, while English to Chinese remains mixed: chrF is competitive, but
BLEU and XCOMET-XL still favor the baseline. This appears partly tied to the Gemma-4 MT backbone,
whose Chinese generations were weaker in our runs; because \AAName{} only
requires a deterministic prompt layout, calibrated heads, and Q/K capture, the
same policy can be ported to other decoder-only LLMs, making translation-focused backbones such as HY-MT-1.5 \citep{hymt1526}, MiLMMT-46 \citep{milmmt4626}, or a Qwen-family MT
backbone, natural next targets for EN$\to$ZH.

\section*{Acknowledgements} 
This work was supported by Czech Operational Program OP JAK, the MSCA CZ project MSCA Fellowships -- Charles University 4, CZ.02.01.01/00/22\_010/0013392, ``LCT''.

\bibliography{references}

\appendix

\section{Additional ASR Analysis}
\label{app:appendix-asr}

The analysis below justifies the source ASR front end used by the cascade
and the recommended source-tail default for future runs.

\paragraph{ASR front-end selection.}
We tested three ASR front ends during development: Qwen3-ASR with the Qwen3
forced aligner, Voxtral Realtime 4B, and a direct Gemma E4B ASR local-agreement
probe. Table~\ref{tab:asr-frontends} summarizes the results on the the MCIF dev set.
Voxtral gives the lowest final WER, but it is not real-time in our serialized
single-GPU loop and its CU-LongYAAL is higher than Qwen3. It also advances the
transcript with a more regular lag, whereas Qwen3 often exposes compact
multi-word updates, which better matches our chunk-synchronous MT policy.
Gemma E4B would be attractive as a single-model substrate, but its WER, prompt
leakage on long talks, and unstable raw latency make it unsuitable here. We
therefore retain Qwen3 forced ASR.

\begin{center}
\centering
\footnotesize
\setlength{\tabcolsep}{3.0pt}
\begin{tabular}{lrrrl}
\toprule
ASR front end & WER $\downarrow$ & CU $\downarrow$ & RTF $\downarrow$ & Decision \\
\midrule
Qwen3 forced & 8.91 & 0.87\,s & 0.34 & keep \\
Voxtral RT 4B & 7.15 & 1.16\,s & 1.34 & reject \\
Gemma E4B LA & 16.69 & $\sim$1.8\,s$^\dagger$ & 0.51 & reject \\
\bottomrule
\end{tabular}
\vspace{-0.6em}
\captionof{table}{\textbf{ASR front-end comparison on 21 MCIF dev talks.}
WER is percent and CU is CU-LongYAAL. $^\dagger$Gemma's raw 0.32\,s LongYAAL
is prompt-leak-contaminated; clipping or discarding negative emissions gives
about 1.8\,s.}
\label{tab:asr-frontends}
\end{center}


\subsection{Offline Cascade Diagnostic}
\label{sec:appendix-offline}
As a development tool that brackets the streaming numbers from above, we also
ran an offline cascade on the same 21 MCIF dev talks: Qwen3-ASR was applied
once to each complete audio file, then Gemma-4 was run in final mode on
sentence-level chunks resegmented as in the streaming evaluation. The full-audio ASR transcripts have 7.40\% weighted
corpus WER against the English reference (90\% audio-bootstrap interval
$[6.27, 8.59]$), so the offline run sees source much closer to the reference
than any online chunk. Differences between offline and streaming output
therefore isolate the cost of the AlignAtt commit policy and the live ASR tail
from the intrinsic quality of the backbone: drops on EN$\to$ZH that already
appear here point to Gemma-4 rather than to the simultaneous policy.

\section{Synchronization Modes}
\label{app:sync}

\begin{figure*}[h]
\centering
\begin{tikzpicture}[
  x=1cm, y=1cm,
  >={Latex[length=1.6mm]},
  asr/.style={
    draw=purple!55!black, rounded corners=2pt, line width=0.55pt,
    fill=purple!7, minimum height=0.52cm, minimum width=1.62cm,
    align=center, font=\scriptsize, text=purple!32!black, inner sep=1.5pt,
  },
  asrsmall/.style={
    draw=purple!55!black, rounded corners=2pt, line width=0.55pt,
    fill=purple!7, minimum height=0.52cm, minimum width=1.44cm,
    align=center, font=\scriptsize, text=purple!32!black, inner sep=1.5pt,
  },
  mt/.style={
    draw=blue!50!black, rounded corners=2pt, line width=0.55pt,
    fill=blue!6, minimum height=0.52cm, align=center,
    font=\scriptsize, text=blue!32!black, inner sep=1.5pt,
  },
  buffer/.style={
    draw=teal!55!black, rounded corners=1.5pt, line width=0.45pt,
    fill=teal!7,
  },
  dispatch/.style={draw=black!22, line width=0.3pt, densely dotted},
  figurebox/.style={draw=black!18, rounded corners=3pt, line width=0.45pt},
  regimeseparator/.style={draw=black!10, line width=0.35pt},
  timeaxis/.style={->, draw=black!50, line width=0.45pt},
  timetick/.style={draw=black!45, line width=0.35pt},
  panelhdr/.style={font=\small\bfseries, text=black!78, anchor=west},
  panelsub/.style={font=\scriptsize\itshape, text=black!52, anchor=west, align=left},
  lanelabel/.style={font=\scriptsize, text=black!60, anchor=east, align=right},
  tick/.style={font=\scriptsize, text=black!50, anchor=north west},
  buflabel/.style={font=\scriptsize\itshape, text=teal!28!black, anchor=center},
]
  \def\xstart{1.55}
  \def\xend{13.75}
  \def\axisy{1.18}

  \draw[figurebox] (0.05, 0.30) rectangle (15.25, 7.35);
  \foreach \x in {1.55, 4.55, 7.55, 10.55, 13.55} {
    \draw[timetick] (\x, {\axisy - 0.05}) -- (\x, {\axisy + 0.05});
  }
  \draw[regimeseparator] (0.55, 5.08) -- (14.80, 5.08);
  \draw[regimeseparator] (0.55, 3.00) -- (14.80, 3.00);

  \node[panelhdr] at (0.35, 7.12) {(a) Full async};
  \node[panelsub, text width=11.6cm] at (2.65, 7.12)
    {: one MT request at a time; relaunch over the latest shared source prefix};

  \node[lanelabel] at (1.20, 6.55) {ASR};
  \foreach \i/\lbl in {0/{$t$}, 1/{$t{+}1$}, 2/{$t{+}2$}, 3/{$t{+}3$}, 4/{$t{+}4$}, 5/{$t{+}5$}, 6/{$t{+}6$}} {
    \node[asrsmall, anchor=west]
      at ({\xstart + \i*1.62}, 6.55) {ASR$(\lbl)$};
  }

  \draw[buffer] (\xstart, 5.98) rectangle (\xend, 6.22);
  \node[buflabel, text width=11.8cm]
    at (7.65, 6.10)
    {shared source prefix \textmd{(whatever ASR has committed at launch)}};
  \fill[teal!60!black] (7.65, 6.24) -- (7.56, 6.35) -- (7.74, 6.35) -- cycle;
  \fill[teal!60!black] (7.65, 5.86) -- (7.56, 5.97) -- (7.74, 5.97) -- cycle;

  \node[lanelabel] at (1.20, 5.56) {MT};
  \node[mt, minimum width=2.35cm, anchor=west] at (2.75, 5.56) {MT};
  \node[mt, minimum width=3.70cm, anchor=west] at (5.10, 5.56) {MT};
  \node[mt, minimum width=1.70cm, anchor=west] at (8.80, 5.56) {MT};
  \node[mt, minimum width=3.00cm, anchor=west] at (10.50, 5.56) {MT};

  \node[panelhdr] at (0.35, 4.74) {(b) Semi async};
  \node[panelsub, text width=11.6cm] at (2.65, 4.74)
    {: fixed ASR cadence; MT$(i)$ dispatches as soon as ASR$(i)$ commits};

  \node[lanelabel] at (1.20, 4.20) {ASR};
  \foreach \x/\lbl in {1.55/{$t$}, 4.55/{$t{+}1$}, 7.55/{$t{+}2$}} {
    \node[asr, anchor=west] at (\x, 4.28) {ASR$(\lbl)$};
  }

  \node[lanelabel] at (1.20, 3.58) {MT};
  \foreach \x in {3.10, 6.10, 9.10} {
    \draw[dispatch] (\x, 4.02) -- (\x, 3.84);
  }
  \node[mt, minimum width=2.90cm, anchor=west] at (3.10, 3.58) {MT$(t)$};
  \node[mt, minimum width=2.90cm, anchor=west] at (6.10, 3.58) {MT$(t{+}1)$};
  \node[mt, minimum width=2.90cm, anchor=west] at (9.10, 3.58) {MT$(t{+}2)$};

  \node[panelhdr] at (0.35, 2.65) {(c) Full sync};
  \node[panelsub, text width=11.6cm] at (2.65, 2.65)
    {: strict ASR/MT alternation; configuration used in our IWSLT~2026 run};

  \node[lanelabel] at (1.20, 2.08) {ASR/MT};
  \node[asr, anchor=west] at (1.55, 2.08) {ASR$(t)$};
  \node[mt,  minimum width=2.00cm, anchor=west] at (3.17, 2.08) {MT$(t)$};
  \node[asr, anchor=west] at (5.45, 2.08) {ASR$(t{+}1)$};
  \node[mt,  minimum width=2.00cm, anchor=west] at (7.07, 2.08) {MT$(t{+}1)$};
  \node[asr, anchor=west] at (9.35, 2.08) {ASR$(t{+}2)$};
  \node[mt,  minimum width=2.00cm, anchor=west] at (10.97, 2.08) {MT$(t{+}2)$};

  \draw[timeaxis] (\xstart, \axisy) -- (\xend, \axisy);
  \node[tick, anchor=north east] at (14.80, {\axisy - 0.08}) {wall-clock time};

\end{tikzpicture}
\captionof{figure}{\textbf{Synchronization regimes for ASR and MT sharing one GPU.}
Regime~(c) is the deployed IWSLT schedule; regimes~(a) and~(b) illustrate
less-blocking alternatives that require asynchronous request handling.}
\label{fig:coord-regimes}
\end{figure*}
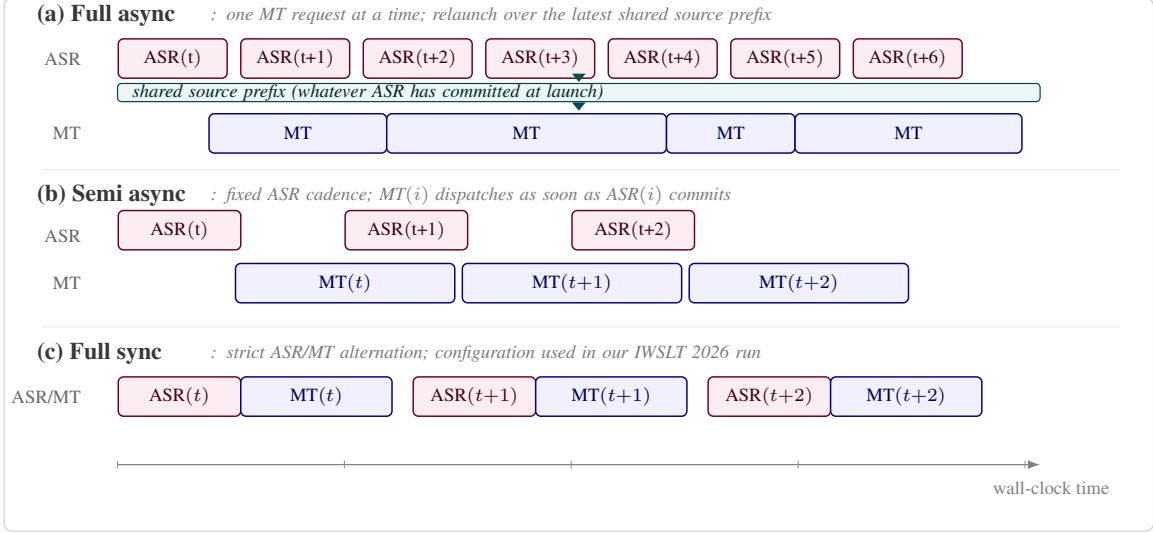

\Cref{fig:coord-regimes} summarizes alternative synchronization modes.

Our cascade loads the ASR and MT models as two separate blocking
\texttt{vllm.LLM} instances on a single GPU, which serializes every audio chunk
into an ASR decode followed by an MT decode
(\Cref{fig:coord-regimes}c). This conservative regime keeps policy
effects separate from scheduler overlap. Alternative pipelined or fully
asynchronous regimes would mainly reduce CA-LongYAAL at fixed CU-LongYAAL, but
would require a non-blocking scheduler and a more
careful treatment of stale MT requests. We leave that scheduler axis to future
work.

\section{MT Alignment-Head Diagnostics}
\label{sec:appendix-alignatt}

See \Cref{fig:mt-alignatt-heads} and \Cref{tab:mt-head-filtering}.

\input{figures/mt_alignment_heads.tex}

\begin{center}
\centering
\footnotesize
\resizebox{\columnwidth}{!}{%
\begin{tabular}{l r r r r}
\toprule
Pair & Top-8 TS & All-336 TS & Gain & Aligned tokens \\
\midrule
EN$\to$DE & 90.40 & 68.49 & \textbf{+21.91} & 11,209 \\
EN$\to$ZH & 93.48 & 65.79 & \textbf{+27.70} & 7,582 \\
EN$\to$IT & 91.90 & 67.42 & \textbf{+24.49} & 12,056 \\
\bottomrule
\end{tabular}
}
\captionof{table}{\textbf{MT head-set filtering ablation on held-out word-aligned dev examples.}
Scores are reported in points ($100\times\mathrm{TS}$) against gold aligned
source tokens.}
\label{tab:mt-head-filtering}
\end{center}

\section{Observer Replay and Qualitative Diagnostics}
\label{sec:appendix-observer}

Below we report the prompt/suffix replay equations, one qualitative probe, and
the numerical parity measurements on the deployed vLLM path.

\subsection{Prompt-space replay details}
After capture, we replay the draft rows against two key blocks: the pre-draft
prompt positions $\mathcal{P}^{(k)}$ and the current draft positions
$\mathcal{D}^{(k)}$, which form the autoregressive suffix. Writing
$\mathbf{B}^{(\ell,h)}_\star := \mathbf{B}^{(\ell)}_\star[h]$ for
head-specific captured tensors, $\gamma^{(\ell)}$ for the attention module's
runtime scaling factor, and $\mathbf{M}^{(\ell,h)}_{\mathrm{prompt}}$ /
$\mathbf{M}^{(\ell,h)}_{\mathrm{draft}}$ for the corresponding
sliding-window and causal/window masks, we first form the two logit blocks
\begin{align}
  \mathbf{P}^{(\ell,h)} &=
  \gamma^{(\ell)}
  \mathbf{B}^{(\ell,h)}_{\mathrm{dQ}}
  \bigl(\mathbf{B}^{(\ell,h)}_{\mathrm{pK}}\bigr)^{\!\top}
  + \mathbf{M}^{(\ell,h)}_{\mathrm{prompt}},
  \label{eq:prompt-logits}\\
  \mathbf{R}^{(\ell,h)} &=
  \gamma^{(\ell)}
  \mathbf{B}^{(\ell,h)}_{\mathrm{dQ}}
  \bigl(\mathbf{B}^{(\ell,h)}_{\mathrm{dK}}\bigr)^{\!\top}
  + \mathbf{M}^{(\ell,h)}_{\mathrm{draft}},
  \label{eq:draft-logits}\\
  \widetilde{\mathbf{A}}^{(\ell,h)}_{\mathcal{D},\,\mathcal{P}\cup\mathcal{D}}
  &=
  \operatorname{softmax}_{\mathrm{row}}
  \!\left(
    \left[\mathbf{P}^{(\ell,h)} \Vert \mathbf{R}^{(\ell,h)}\right]
  \right).
  \label{eq:vllm-softmax}
\end{align}
The row-wise softmax is taken over the concatenated prompt and draft columns,
so source positions compete with all other causal context exactly as in the
deployed attention row. Restricting
$\widetilde{\mathbf{A}}^{(\ell,h)}_{\mathcal{D},\,\mathcal{P}\cup\mathcal{D}}$
to the source columns $\phi^{(k)}(s)$ recovers the policy-visible block of
Eq.~\eqref{eq:mt-qk-fast}. The replay cost therefore scales with the selected
head set and the short draft length, not with the full $LHn^2$ attention
tensor.

\subsection{Qualitative reconstruction example}
Figure~\ref{fig:mt-selective-reconstruction} shows the full reconstruction
stack on an EN$\to$DE text-only MT probe: the exact prompt partition seen by
the decoder, word-level aggregation of the replayed rows for readability, and
the four-way provenance accounting for every drafted word.

\input{figures/mt_selective_reconstruction.tex}

The green band in Figure~\ref{fig:mt-selective-reconstruction} is previously
committed target text reused as causal context; it is not itself the object of
the gate. The orange words are the current draft scanned left-to-right by the
policy. We allow $\hat{s}_t$ to move backward within a live draft when the
decoder revisits earlier source material under reordering, which is why the
residual prompt and suffix masses remain explicit in the decoder-only
formulation.

\subsection{Numerical parity}
Let $\mathbf{A}^{(\mathrm{TF})}$ denote the reference attention tensor produced
by a Transformers reference forward on the identical prompt, and let
$\Delta\mathbf{A} =
\widetilde{\mathbf{A}} - \mathbf{A}^{(\mathrm{TF})}$. On a curated parity set,
the reference and deployed paths make bit-identical acceptance decisions and
satisfy
\begin{equation}
  \begin{aligned}
    \|\Delta\mathbf{A}\|_\infty &\leq 1.2 \times 10^{-2}, \\
    \|\Delta\mathbf{A}\|_1 / (nL_k) &\leq 4 \times 10^{-4}.
  \end{aligned}
  \label{eq:parity}
\end{equation}
\end{document}